\pdfoutput=1

\documentclass{article}
\usepackage{denglab-qwen-style}

\usepackage{url}
\usepackage{booktabs}
\usepackage{hyperref}
\hypersetup{
  pdftitle={Wave Forcing: Towards Speed-of-Light Streaming Video Generation},
  pdfauthor={Yijie Jin, Yuxuan Liu, Jiajun Xu, Ye Huang, Chuzhi Wang, Jiayang Li, Chenling Hong, Xu Wang, Zhijie Deng}
}
\renewcommand{\ghlink}{https://github.com/SJTU-DENG-Lab/WaveForcing}
\newcommand{\projectlink}{https://sjtu-deng-lab.github.io/WaveForcing/}
\renewcommand{\hflink}{https://huggingface.co/SJTU-DENG-Lab/WaveForcing-1.3B}
\usepackage{graphicx}
\usepackage{fontawesome5}
\usepackage{microtype}
\usepackage{multicol}
\usepackage{xcolor}
\usepackage{multirow}
\usepackage{lipsum}
\usepackage{enumitem}
\usepackage{pifont}
\usepackage{colortbl}
\usepackage{tabularx}
\usepackage{caption}
\usepackage{float}
\usepackage{placeins}
\usepackage{wrapfig}
\usepackage{needspace}
\usepackage{amsmath,amsfonts,amssymb,bbm}
\usepackage{amsthm}
\usepackage{algorithm}
\usepackage{algpseudocode}
\usepackage{subfig}
\usepackage{array}
\usepackage{tikz}
\usetikzlibrary{arrows.meta,positioning,calc,patterns,decorations.pathreplacing}
\definecolor{waveblue}{HTML}{3568A4}
\definecolor{waveteal}{HTML}{218477}
\definecolor{waveorange}{HTML}{BC7131}
\definecolor{wavegray}{HTML}{52606D}
\definecolor{tableteal}{HTML}{427C74}
\definecolor{tablecoral}{HTML}{A45C53}
\definecolor{tableamber}{HTML}{946C35}
\definecolor{tabletealfill}{HTML}{DCEDEA}
\definecolor{tablecoralfill}{HTML}{EFD6D2}
\colorlet{tablebestbg}{tabletealfill}
\colorlet{tableissuebg}{tablecoralfill}

\DeclareRobustCommand{\tabhead}[1]{\textcolor{black}{#1}}
\DeclareRobustCommand{\tabbest}[1]{%
  \cellcolor{tablebestbg}\textcolor{tableteal}{\textbf{#1}}}
\DeclareRobustCommand{\tabbestmath}[1]{%
  \cellcolor{tablebestbg}\textcolor{tableteal}{\mathbf{#1}}}
\DeclareRobustCommand{\tabissue}[1]{%
  \cellcolor{tableissuebg}%
  \textcolor{tablecoral}{#1}}
\DeclareRobustCommand{\tabbestinline}[1]{%
  \begingroup\setlength{\fboxsep}{1.5pt}%
  \colorbox{tablebestbg}{\textcolor{tableteal}{\textbf{#1}}}\endgroup}
\DeclareRobustCommand{\tabissueinline}[1]{%
  \begingroup\setlength{\fboxsep}{1.5pt}%
  \colorbox{tableissuebg}{\textcolor{tablecoral}{#1}}\endgroup}
\DeclareRobustCommand{\tabgain}[1]{#1}
\DeclareRobustCommand{\tabloss}[1]{#1}
\DeclareRobustCommand{\tabdiagmark}{%
  \textcolor{tableamber}{\ensuremath{^{\dagger}}}}

\newtheorem{proposition}{Proposition}

\newcommand*\samethanks[1][\value{footnote}]{\footnotemark[1]}

\makeatletter
\def\@fnsymbol#1{\ensuremath{\ifcase#1\or \dagger \or  \ddagger\or
   \mathsection\or  \text{*}\or \mathparagraph \or  \| \or **\or \dagger\dagger
   \or \ddagger\ddagger \else\@ctrerr\fi}}
\makeatother
\renewcommand{\thefootnote}{\fnsymbol{footnote}}

\title{Wave Forcing: Towards Speed-of-Light\\Streaming Video Generation}

\author{
    \textbf{Yijie Jin}$^{1}$\footnotemark[1] \quad
    \textbf{Yuxuan Liu}$^{1}$\footnotemark[1] \quad
    \textbf{Jiajun Xu}$^{1,2}$\footnotemark[1] \quad
    \textbf{Ye Huang}$^{3}$\footnotemark[1] \quad
    \textbf{Chuzhi Wang}$^{1}$ \\[0.5mm]
    \textbf{Jiayang Li}$^{4}$ \quad
    \textbf{Ruiyang Chen}$^{1}$ \quad
    \textbf{Chenling Hong}$^{5}$ \quad
    \textbf{Xu Wang}$^{1}$ \quad
    \textbf{Zhijie Deng}$^{1}$\footnotemark[2] \\[1.5mm]
    $^{1}$DENG Lab, Shanghai Jiao Tong University \\
    $^{2}$Xi'an Jiaotong University \quad
    $^{3}$Peking University \\
    $^{4}$East China Normal University \quad
    $^{5}$National Tsing Hua University \\[1mm]
    \texttt{\detokenize{{drewjin0827, liuyx2005, zhijied}@sjtu.edu.cn}} \\
    \texttt{mikanaffine@stu.xjtu.edu.cn, huangye26@stu.pku.edu.cn} \\[1.5mm]
    \href{\ghlink}{\github\enspace\texttt{GitHub}} \quad
    \href{\projectlink}{\faGlobe\enspace\texttt{Project Page}} \quad
    \href{\hflink}{\huggingface\enspace\texttt{Hugging Face}}
}

\begin{document}
\maketitle
\suppressfloats[t]

\footnotetext[1]{Equal contribution.}
\footnotetext[2]{Corresponding author.}
\renewcommand{\thefootnote}{\arabic{footnote}}
\setcounter{footnote}{0}

\begin{abstract}
Real-time long-video generation requires continuous output with consistent appearance and natural motion.
Existing chunk-wise autoregressive methods face challenges in both generation efficiency and long-video quality: serialized denoising and history-cache refresh require two forward passes even with one-step denoising, and longer rollouts can degrade quality and cause camera drift.
We propose \textbf{Wave Forcing (WF)}, which co-designs generation, training, and execution for continuous video output with multi-step denoising.
WF uses chunk-wise causal attention to access earlier, cleaner chunks before their denoising is complete, and distributes staggered denoising and cache-refresh stages across full-model replicas to achieve \textbf{1F1C}: one output chunk per concurrent forward round after pipeline fill.
To adapt the model to this continuous generation process, we develop a three-stage training procedure that successively introduces mixed-noise denoising, motion supervision, and causal adaptation using self-generated history.
To execute this process efficiently, \textbf{Wave Runtime (WaveRT)} uses \textbf{Wave Parallelism (WP)} for early K/V delivery and communication--computation overlap, together with optimized kernels and pipelined VAE decoding.
Experiments on Wan2.1-T2V-1.3B show that 5-step WF maintains overall quality comparable to Rolling Forcing while achieving higher Dynamic Degree on both short and long videos.
Our system efficiency experiments show that 4-step WF achieves \textbf{147.6 DiT FPS} and \textbf{134.8 end-to-end FPS} on eight H200 GPUs, yielding $\mathbf{7.07\times}$ the single-GPU end-to-end throughput.
In DiT-only tests with matched SageAttention+W8A8 backends, denoising steps, and GPU counts, WP delivers up to $\mathbf{2.53\times}$ the throughput of the faster of the Ulysses and Ring context-parallel baselines across six model scales (1.3B--14B).
\end{abstract}

\section{Introduction}
\label{sec:intro}

\begin{figure}[t]
\centering
\includegraphics[width=\linewidth]{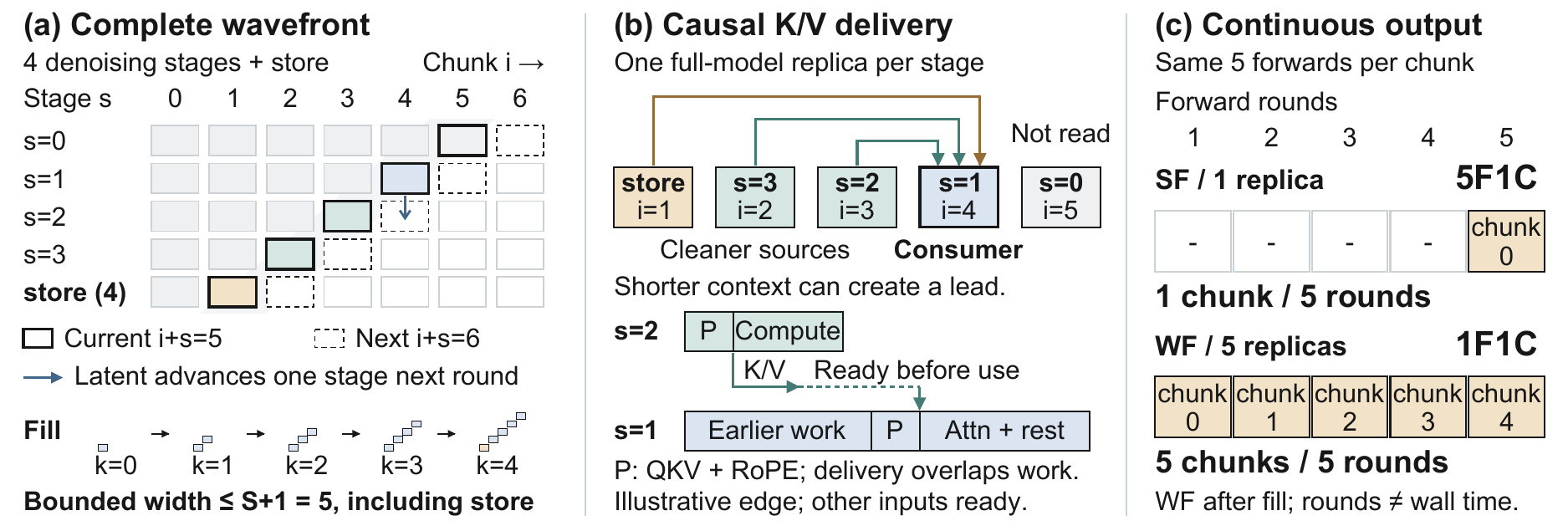}
\caption{Causal wavefront and continuous output ($S=4$). (a) Denoising and store advance on anti-diagonals, with dashed cells marking the next round. (b) One later-layer K/V edge arrives before use. Other edges and initial waits are omitted. (c) Both schedules retain four denoising forwards plus store per chunk. Over five rounds, one SF replica emits one chunk, whereas five WF replicas emit five after fill. Wall-clock gains depend on stage costs and overlap.}
\label{fig:wave-cadence}
\end{figure}

Real-time streaming video generation must sustain playable output with consistent appearance and natural motion.
Chunk-wise autoregressive methods such as Self Forcing (SF) and Causal Forcing (CF) denoise each chunk conditioned on preceding content~\citep{huang2025selfforcing,zhu2026causalforcing}.
Under the clean-history caching convention adopted in this work, an additional forward pass, termed \emph{store}, refreshes the key/value (K/V) history from the clean result.
Sequential execution places denoising and store between consecutive outputs, yielding \textbf{NF1C (N Forward Rounds, 1 Chunk Output)}.
Thus, even one-step denoising requires two forward passes when store is included.

We target \textbf{1F1C}: one completed chunk per concurrent forward round after pipeline fill, while retaining every denoising update.
Inspired by wavefront scheduling~\citep{jin2016hybridwavefront}, earlier chunks execute later denoising stages while newer chunks execute earlier ones.
These staggered tasks form a moving \emph{wavefront} (Fig.~\ref{fig:wave-cadence}a).
Assigning a full-model replica to each denoising stage and the store stage creates a bounded-width pipeline (Fig.~\ref{fig:wave-cadence}c).

To advance chunks before their predecessors finish, we use \emph{chunk-wise causal context}: full attention within each chunk, cached history, and live context from earlier, cleaner chunks, with no dependence on later chunks.
This directional conditioning accommodates differences in stage-wise progress and enables overlapped execution (Fig.~\ref{fig:wave-cadence}b).
Our staged training builds on mixed-noise joint denoising~\citep{liu2025rollingforcing}, uses teacher videos to supervise motion and framing, and adapts the model to causal inference with self-generated history.
This generation structure and its training recipe together constitute \textbf{Wave~Forcing~(WF)}.

To turn this scheduling potential into shorter output intervals, \textbf{Wave Runtime (WaveRT)} implements \textbf{Wave Parallelism (WP)}.
Following GPU software-pipelining principles~\citep{nvidia2026ampere,nvidia2026cutlass}, WP exploits differences in stage-wise progress for early layer-wise K/V delivery and communication--computation overlap, reducing exposed dependency waits.
Data placement, kernel optimization, and load-balanced VAE pipeline parallelism complete the execution path from concurrent denoising to \mbox{sustained video output}.

On Wan2.1-T2V-1.3B~\citep{wan2025wan}, 5-step WF achieves overall quality comparable to 5-step Rolling Forcing (RF), with VBench Total scores of 81.64 versus 81.15 and VBench-Long Quality scores of 81.26 versus 81.18~\citep{huang2024vbench,huang2025vbenchplus}.
Dynamic Degree increases from 51.39 to 94.44 on short videos and from 36.63 to 91.45 on 100-second videos.
These gains are accompanied by better Subject Preservation (97.10 versus RF's 92.75) while maintaining low Camera Drift \mbox{(0.37 versus RF's 0.17)}.

In terms of system performance, 4-step WF with the BF16/FA3 backend achieves \textbf{147.6 DiT FPS} and \textbf{134.8 end-to-end FPS} on eight H200 GPUs, yielding a $\mathbf{7.07\times}$ end-to-end speedup over the 19.06-FPS single-GPU baseline (Table~\ref{tab:runtime-core}).
At this denoising speed, residual output-path overhead remains visible even with three-stage VAE pipelining, as discussed in Section~\ref{sec:vae-limitations}.
Across six model scales from 1.3B to 14B, WP outperforms the faster context-parallel (CP) baseline at five scales under BF16 and all six under Sage+W8A8, achieving $\mathbf{1.025\text{--}2.530\times}$ its throughput under Sage+W8A8.
System ablations and profiling further analyze how communication overlap, K/V placement, compute backends, and VAE stage balance affect runtime performance.

Our contributions are threefold:
\begin{itemize}
    \item \textbf{Causal wavefront generation.}
    We introduce Wave Forcing, allowing newer chunks to condition on earlier, cleaner chunks before their denoising completes. By pipelining denoising and clean-history cache refresh across full-model replicas, WF enables one completed chunk per concurrent forward round (1F1C) after pipeline fill while retaining all denoising updates.

    \item \textbf{Training for continuous generation.}
    We develop a staged training recipe combining mixed-noise initialization, teacher-guided motion supervision, and causal adaptation with self-generated history. This adapts the model to wavefront inference, enabling strong motion and subject preservation over long rollouts at overall quality comparable to Rolling Forcing.

    \item \textbf{Efficient streaming execution.}
    We build Wave Runtime to realize wavefront concurrency through early layer-wise K/V delivery, communication--computation overlap, and direct placement into attention buffers. With optimized local computation and load-balanced VAE decoding, the runtime translates stage concurrency into sustained end-to-end throughput, with efficiency validated across 1.3B--14B models.
\end{itemize}
\section{Preliminaries}
\label{sec:background}

\paragraph{Chunk denoising and store.}
A VAE decodes latent-frame chunks into video~\citep{wan2025wan}. Following SF~\citep{huang2025selfforcing}, chunk $i$ undergoes $S$ history-conditioned denoising updates, from noise $Z_i^{(0)}$ to clean output $B_i=Z_i^{(S)}$. At noise timesteps $\nu_0>\cdots>\nu_S=0$,
\begin{equation}
 Z_i^{(s+1)}
 =\Psi_{\nu_{s+1}}\!\left(
 G_\theta(Z_i^{(s)},\nu_s\mid\mathcal C_{i,s}),\epsilon_{i,s}\right),
 \qquad 0\leq s<S,
 \label{eq:stage-update}
\end{equation}
$G_\theta$ predicts clean latents, which $\Psi_\nu$ re-noises using draw $\epsilon_{i,s}$, with $\Psi_0(x,\epsilon)=x$. $\mathcal C_{i,s}$ denotes layer-wise K/V context, with text conditioning omitted for brevity. Denoising K/V represents the noisy input, so our clean-history convention adds \emph{store}: a full-model forward on $B_i$ that retains K/V and discards the prediction. Sequential generation thus uses $N=S+1$ forwards per chunk (NF1C).

\paragraph{Wavefront states and execution rounds.}
At filled runtime tick $k$, the denoising inputs are
\begin{equation}
 \mathcal F_k=
 \{Z_k^{(0)},Z_{k-1}^{(1)},\ldots,Z_{k-S+1}^{(S-1)}\}.
 \label{eq:frontier}
\end{equation}
Adding store at $s=S$ gives $\mathcal W_k=\mathcal F_k\cup\{B_{k-S}\}$, of width at most $S+1$ excluding anchor/history caches. Each round advances active stages and admits fresh noise. With stage-owned replicas, this schedule permits 1F1C after fill without changing per-chunk work.

\section{Wave Forcing}
\label{sec:wf}

We introduce \textbf{Wave Forcing (WF)} to advance successive chunks before their predecessors finish denoising. WF couples a causal wavefront---newer, noisier chunks read earlier, cleaner states, but not vice versa---with training for this mixed-noise context and self-generated history.

\subsection{A bounded-width causal wavefront}

Cell $(i,s)$, $s<S$, in Fig.~\ref{fig:wave-cadence}a updates $Z_i^{(s)}$ to $Z_i^{(s+1)}$ in round
\begin{equation}
 k=i+s,\qquad i=k-s.
 \label{eq:wave-coordinate}
\end{equation}
Store at $(k-S,S)$ refreshes $B_{k-S}$'s K/V. For $N_{\mathrm{chunk}}$ requested chunks, active denoising stages are $\mathcal A_k=\{s:0\leq s<S,\ 0\leq k-s<N_{\mathrm{chunk}}\}$. Store is active when $0\leq k-S<N_{\mathrm{chunk}}$.

WF retains full attention within each chunk and uses \emph{chunk-wise causal attention} across video time. Query slot $s$ sees K/V slot $j$ iff
\begin{equation}
 M_{\mathrm{WF}}(s,j)=\mathbf 1[j\geq s],
 \qquad s,j\in\mathcal A_k.
 \label{eq:wf-mask}
\end{equation}
In chunk order, $i'=k-j\leq i=k-s$ gives a lower-triangular mask over self and \emph{all} \mbox{cleaner predecessors}.

At Transformer layer $\ell\in\{1,\ldots,L\}$, the ordered context is
\begin{equation}
 \mathcal C_{k,s}^{\mathrm{WF},\ell}
 = A_k^\ell\oplus H_k^\ell\oplus U_k^\ell
   \oplus \Bigl[\bigoplus_{\substack{j\in\mathcal A_k\\j\geq s}}
                 (K_{k,j}^\ell,V_{k,j}^\ell)\Bigr],
 \label{eq:wf-context}
\end{equation}
Here $\oplus$ concatenates initial-chunk anchor $A_k^\ell$, bounded clean history $H_k^\ell$, live store $U_k^\ell$, and denoising K/V in increasing $j$ (self first). Store appears once, is empty when inactive, and enters history after the round. Appendix~\ref{sec:runtime-appendix} details fill/drain and caching.

This removes Rolling Forcing's (RF) reverse refinement~\citep{liu2025rollingforcing}. Without post-training, short-video Quality and Dynamic Degree fall in our evaluation (Sec.~\ref{sec:quality}), motivating adaptation.

\subsection{Training for the deployment graph}
\label{sec:training}

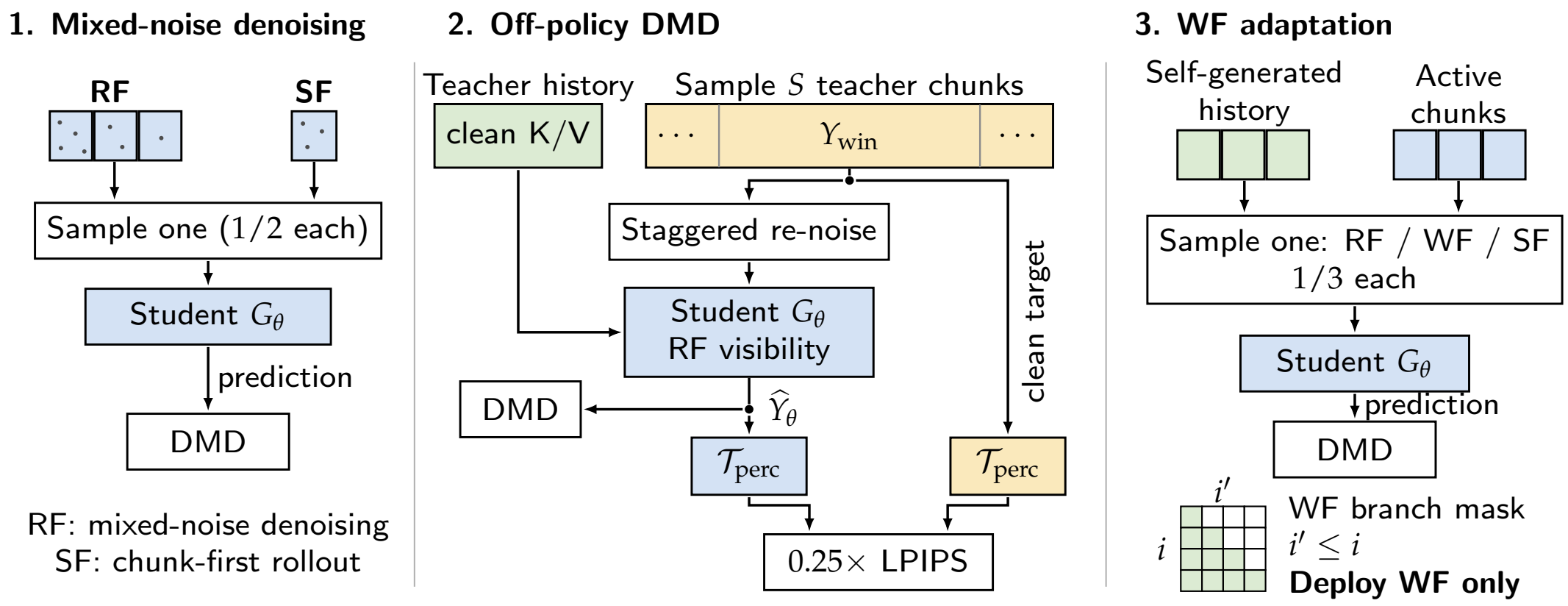
\begin{figure}[t]
\centering
\resizebox{0.88\textwidth}{!}{
\begin{tikzpicture}[x=1cm,y=1cm,>={Latex[length=1.2mm,width=.9mm]},
  font=\sffamily\fontsize{8}{9.5}\selectfont,
  op/.style={draw=black,line width=.55pt,rectangle,align=center,
    minimum height=.5cm,inner sep=3pt},
  wire/.style={draw=black,line width=.65pt},
  flow/.style={wire,->,shorten <=.35pt,shorten >=.35pt},
  junction/.style={circle,fill=black,inner sep=0pt,minimum size=2.2pt},
  note/.style={align=center,inner sep=0pt},
  heading/.style={anchor=west,font=\sffamily\fontsize{8.5}{10}\selectfont\bfseries,
    inner sep=0pt}]
\definecolor{trainblue}{HTML}{D4E2F4}
\definecolor{traingold}{HTML}{FAE9BC}
\definecolor{traingreen}{HTML}{DCEBD3}
\draw[black!30,line width=.4pt] (3.65,-.32) -- (3.65,-5.02);
\draw[black!30,line width=.4pt] (9.85,-.32) -- (9.85,-5.02);
\node[heading] at (0,0) {1. Mixed-noise denoising};
\node[heading] at (3.94,0) {2. Off-policy DMD};
\node[heading] at (10.1,0) {3. WF adaptation};

\node[note] at (.92,-.58) {\textbf{RF}};
\node[note] at (2.75,-.58) {\textbf{SF}};
\foreach \x in {.38,.78,1.18} {
  \draw[fill=trainblue,line width=.55pt] (\x,-1.20) rectangle ++(.38,.44);
}
\foreach \x/\y in {.47/-.86,.61/-.96,.48/-1.09,.69/-1.12,.92/-.9,1.03/-1.08,1.38/-1.0} {
  \fill[black!70] (\x,\y) circle[radius=.55pt];
}
\draw[fill=trainblue,line width=.55pt] (2.55,-1.20) rectangle ++(.4,.44);
\foreach \x/\y in {2.65/-.87,2.82/-.94,2.69/-1.08} {
  \fill[black!70] (\x,\y) circle[radius=.55pt];
}
\node[op,minimum width=3.05cm] (sample1) at (1.8,-1.83) {Sample one ($1/2$ each)};
\coordinate (rf1out) at (.96,-1.20);
\coordinate (sf1out) at (2.75,-1.20);
\draw[flow] (rf1out) -- (rf1out |- sample1.north);
\draw[flow] (sf1out) -- (sf1out |- sample1.north);
\node[op,fill=trainblue,minimum width=2.2cm] (s1) at (1.8,-2.6) {Student $G_\theta$};
\draw[flow] (sample1.south) -- (s1.north);
\node[op,minimum width=1.45cm] (dmd1) at (1.8,-3.72) {DMD};
\draw[flow] (s1.south) -- node[note,right,xshift=2pt] {prediction} (dmd1.north);
\node[note] at (1.8,-4.62) {RF: mixed-noise denoising\\SF: chunk-first rollout};

\node[note] at (4.67,-.54) {Teacher history};
\node[op,fill=traingreen,minimum width=1.06cm,minimum height=.57cm] (hist2)
  at (4.58,-.98) {clean K/V};
\node[note] at (7.55,-.54) {Sample $S$ teacher chunks};
\node[op,fill=traingold,minimum width=3.65cm,minimum height=.57cm] (win2)
  at (7.55,-.98) {$Y_{\mathrm{win}}$};
\draw[black!45,line width=.45pt] (6.38,-1.26) -- (6.38,-.70);
\draw[black!45,line width=.45pt] (8.72,-1.26) -- (8.72,-.70);
\node[note] at (6.02,-.98) {$\cdots$};
\node[note] at (9.08,-.98) {$\cdots$};
\node[op,minimum width=2.23cm] (noise2) at (6.65,-1.84) {Staggered re-noise};
\node[junction] (teacher2split) at (7.55,-1.38) {};
\draw[wire] (win2.south) -- (teacher2split.north);
\draw[flow] (teacher2split.west) -| (noise2.north);
\node[op,fill=trainblue,minimum width=2.23cm] (s2) at (6.65,-2.74) {Student $G_\theta$\\RF visibility};
\draw[flow] (noise2.south) -- (s2.north);
\draw[flow] (hist2.south) |- (s2.west);
\node[junction] (pred2) at (6.65,-3.43) {};
\draw[wire] (s2.south) -- (pred2.north);
\node[note,anchor=west] at (6.81,-3.42) {$\widehat Y_\theta$};
\node[op,minimum width=1.08cm] (dmd2) at (4.6,-3.43) {DMD};
\draw[flow] (pred2.west) -- (dmd2.east);
\node[op,fill=trainblue,minimum width=1.03cm] (perc2) at (6.65,-3.94) {$\mathcal T_{\mathrm{perc}}$};
\node[op,fill=traingold,minimum width=1.03cm] (target2) at (8.97,-3.94) {$\mathcal T_{\mathrm{perc}}$};
\draw[flow] (pred2.south) -- (perc2.north);
\draw[flow] (teacher2split.east) -| (target2.north);
\node[note,rotate=90] at (9.22,-2.64) {clean target};
\node[op,minimum width=2.05cm] (lpips2) at (7.81,-4.80) {$0.25\times$ LPIPS};
\coordinate (lpips2pred) at ($(lpips2.north)+(-.62,0)$);
\coordinate (lpips2target) at ($(lpips2.north)+(.62,0)$);
\draw[flow] (perc2.south) -- (6.65,-4.29) -| (lpips2pred);
\draw[flow] (target2.south) -- (8.97,-4.29) -| (lpips2target);

\node[note] at (11.09,-.60) {Self-generated\\history};
\node[note] at (13.01,-.60) {Active\\chunks};
\foreach \x in {10.49,10.89,11.29} {
  \draw[fill=traingreen,line width=.55pt] (\x,-1.37) rectangle ++(.38,.44);
}
\foreach \x in {12.43,12.83,13.23} {
  \draw[fill=trainblue,line width=.55pt] (\x,-1.37) rectangle ++(.38,.44);
}
\node[op,minimum width=3.35cm] (sample3) at (12.08,-2.09)
  {Sample one: RF / WF / SF\\$1/3$ each};
\coordinate (hist3out) at (11.09,-1.37);
\coordinate (active3out) at (13.01,-1.37);
\draw[flow] (hist3out) -- (hist3out |- sample3.north);
\draw[flow] (active3out) -- (active3out |- sample3.north);
\node[op,fill=trainblue,minimum width=2.05cm] (s3) at (12.08,-3.02) {Student $G_\theta$};
\draw[flow] (sample3.south) -- (s3.north);
\node[op,minimum width=1.45cm] (dmd3) at (12.08,-3.79) {DMD};
\draw[flow] (s3.south) -- node[note,right,xshift=2pt] {prediction} (dmd3.north);
\foreach \row in {0,...,3} {
  \foreach \col in {0,...,3} {
    \ifnum\col>\row
      \def\cellfill{white}
    \else
      \def\cellfill{traingreen}
    \fi
    \draw[fill=\cellfill,line width=.4pt]
      ({10.52+.19*\col},{-4.3-.19*\row}) rectangle ++(.19,-.19);
  }
}
\node[note] at (10.90,-4.15) {$i'$};
\node[note] at (10.35,-4.68) {$i$};
\node[note,anchor=west,align=left] at (11.48,-4.67)
  {WF branch mask\\$i'\leq i$\\\textbf{Deploy WF only}};
\end{tikzpicture}}
\caption{Training flow. Stage 1 applies DMD to one sampled RF or SF rollout. Stage 2 re-noises teacher chunks and applies DMD and LPIPS to the same prediction $\widehat Y_\theta$. Stage 3 samples one self-generated RF, WF, or SF rollout for on-policy DMD.}
\label{fig:training-stages}
\end{figure}

The causal mask removes access to later, noisier active chunks, changing RF's learned conditioning. We adapt Wan2.1-T2V-1.3B~\citep{wan2025wan} through mixed-noise initialization, teacher guidance, and on-policy training (Fig.~\ref{fig:training-stages}). The 4-step student starts from SF's ODE initialization, whereas the 5-step student reuses released RF weights and skips Stage~1.

\paragraph{Stage~1: mixed-noise initialization.}
We adopt RF's mixed-noise joint-denoising setup~\citep{liu2025rollingforcing}, using bidirectional attention across chunks at different noise levels. We sample either an RF rollout or an SF rollout with equal probability and train the student with distribution matching distillation (DMD)~\citep{yin2024dmd}. SF rollouts help stabilize motion during initialization.

\paragraph{Stage~2: off-policy DMD.}
Following LPM~1.0~\citep{zeng2026lpm}, we re-noise consecutive teacher chunks $Y_{\mathrm{win}}$ at staggered timesteps, with preceding teacher chunks as clean K/V history. Retaining Stage~1's bidirectional attention, this supplies stronger motion and subject guidance (Sec.~\ref{sec:training-study}). DMD and LPIPS~\citep{zhang2018lpips} supervise the same text-conditioned prediction $\widehat Y_\theta$:
\begin{equation}
 \mathcal L_2
 =\mathbb E\!\left[
   \mathcal D_{\mathrm{DMD}}(\widehat Y_\theta;\mathbf c)
   +\lambda_{\mathrm p}\operatorname{LPIPS}\!\left(
      \mathcal T_{\mathrm{perc}}(\widehat Y_\theta),
      \mathcal T_{\mathrm{perc}}(Y_{\mathrm{win}})\right)\right],
 \qquad \lambda_{\mathrm p}=0.25,
 \label{eq:training-stage2}
\end{equation}
Here $\mathcal D_{\mathrm{DMD}}$ is the DMD objective, and $\mathcal T_{\mathrm{perc}}$ decodes downsampled latents to sampled RGB frames.

\paragraph{Stage~3: causal context and self-generated history.}
Each iteration uniformly samples a self-generated RF, WF, or SF rollout for on-policy DMD. WF uses Eq.~\eqref{eq:wf-context}, including store-refreshed history from its own outputs. RF and SF provide complementary training distributions, but only WF is deployed. For branch objective $\mathcal L_b$,
\begin{equation}
 \mathcal L_3=\mathbb E_{b\sim\mathrm{Unif}\{\mathrm{RF},\mathrm{WF},\mathrm{SF}\}}
                  [\mathcal L_b]
             =\tfrac13\mathcal L_{\mathrm{RF}}
              +\tfrac13\mathcal L_{\mathrm{WF}}
              +\tfrac13\mathcal L_{\mathrm{SF}}.
 \label{eq:training-stage3}
\end{equation}
Appendix~\ref{sec:training-appendix} gives stage transitions and implementation details. Together, these stages adapt the model to causal wavefront inference: chunks denoise using cleaner, still-evolving predecessors and self-generated history.

\FloatBarrier
\suppressfloats[t]
\section{Wave Runtime}
\label{sec:wp}
\label{sec:system}

WF exposes stage concurrency, but K/V dependencies can still stall attention. \textbf{Wave Runtime (WaveRT)} realizes \textbf{Wave Parallelism (WP)} by publishing K/V early from stage-owned replicas and overlapping delivery with computation when cleaner producers run ahead. We use SGLang Diffusion's model implementation as the core execution backend~\citep{zheng2024sglang} of WaveRT.

\begin{figure}[t]
\centering
\includegraphics[width=\linewidth]{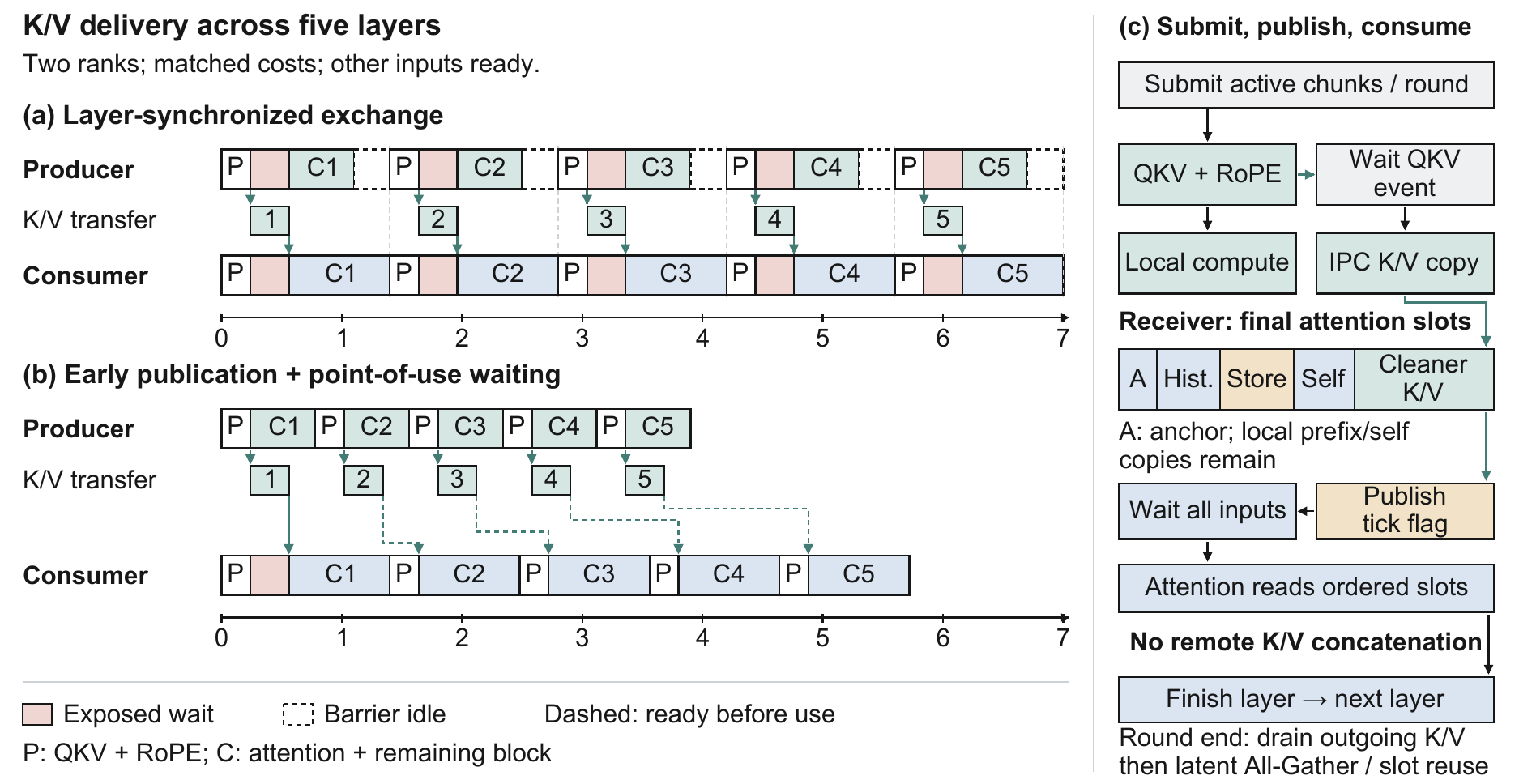}
\caption{Hiding K/V delivery with identical schematic operation costs. (a) Layer barriers expose communication. (b) Early publication hides later waits, not dependencies or the initial wait. (c) Direct writes to receiver slots precede readiness signals and consumption.}
\label{fig:wave-overlap}
\end{figure}

\subsection{Stage placement and dependencies}

Each denoising stage and store owns a full-model GPU replica. Denoising rank $s$ updates $Z_{k-s}^{(s)}$ and passes it to rank $s+1$ for the next round (Fig.~\ref{fig:wave-cadence}). Within a round, let $P_{s,\ell}$ produce Q/K/V and RoPE, and $C_{s,\ell}$ perform attention and the remaining block:
\begin{align}
 P_{s,\ell}\rightarrow C_{s,\ell},\qquad
 C_{s,\ell}\rightarrow P_{s,\ell+1}\quad(\ell<L),
 \label{eq:local-edge}\\
 P_{j,\ell}\rightarrow C_{s,\ell},\qquad j>s,\quad s,j\in\mathcal A_k.
 \label{eq:cross-edge}
\end{align}
At each layer, local projection can proceed without waiting for remote K/V, but attention must wait until store and all legal cleaner ranks, including nonadjacent ranks, have supplied their K/V. RF additionally depends on noisier ranks, introducing reverse edges that WF removes through its causal mask. The synchronized reference enforces these dependencies with a per-layer all-rank K/V gather.

\subsection{Producer lead and asynchronous delivery}

In a filled round, rank $s$ reads $N_C$ query tokens and $N_s$ keys, where
\begin{equation}
 N_s=N_{\mathrm{prefix}}+(S-s)N_C.
 \label{eq:context-skew}
\end{equation}
Here $N_C$ is tokens per chunk and $N_{\mathrm{prefix}}$ covers anchor, history, and store. Cleaner ranks read fewer keys, permitting later-layer producer lead. WaveRT publishes K/V after QKV projection and RoPE, then checks readiness immediately before attention (Fig.~\ref{fig:wave-overlap}).

On a common time base, edge $e$ has payload-ready time $t_e^{\mathrm{ready}}$, consumer local-ready time $t_e^{\mathrm{need}}$ before remote waits, and delivery cost $\tau_e$ including queueing, transfer, and signaling:
\begin{equation}
 \Delta_e=t_e^{\mathrm{need}}-t_e^{\mathrm{ready}},\qquad
 w_e=\max(0,\tau_e-\Delta_e).
 \label{eq:slack}
\end{equation}
A consumer's exposed wait is the maximum $w_e$ over required inputs. Thus, delivery is fully hidden only when every $\tau_e\leq\Delta_e$ (Appendix~\ref{sec:timing-contract}).

Following software-pipeline principles~\citep{nvidia2026ampere}, \emph{paged} delivery writes K/V directly into ordered attention slots and signals readiness before use, avoiding remote concatenation. When weights, noise draws, positions, and local operators are held fixed, attention consumes the same K/V values as in synchronized execution (Appendix~\ref{sec:semantic-correctness}).
Appendix~\ref{sec:runtime-appendix} details buffer lifetimes and readiness signaling.

\subsection{Local computation and streaming decode}

\begin{figure}[t]
\centering
\includegraphics[width=\linewidth]{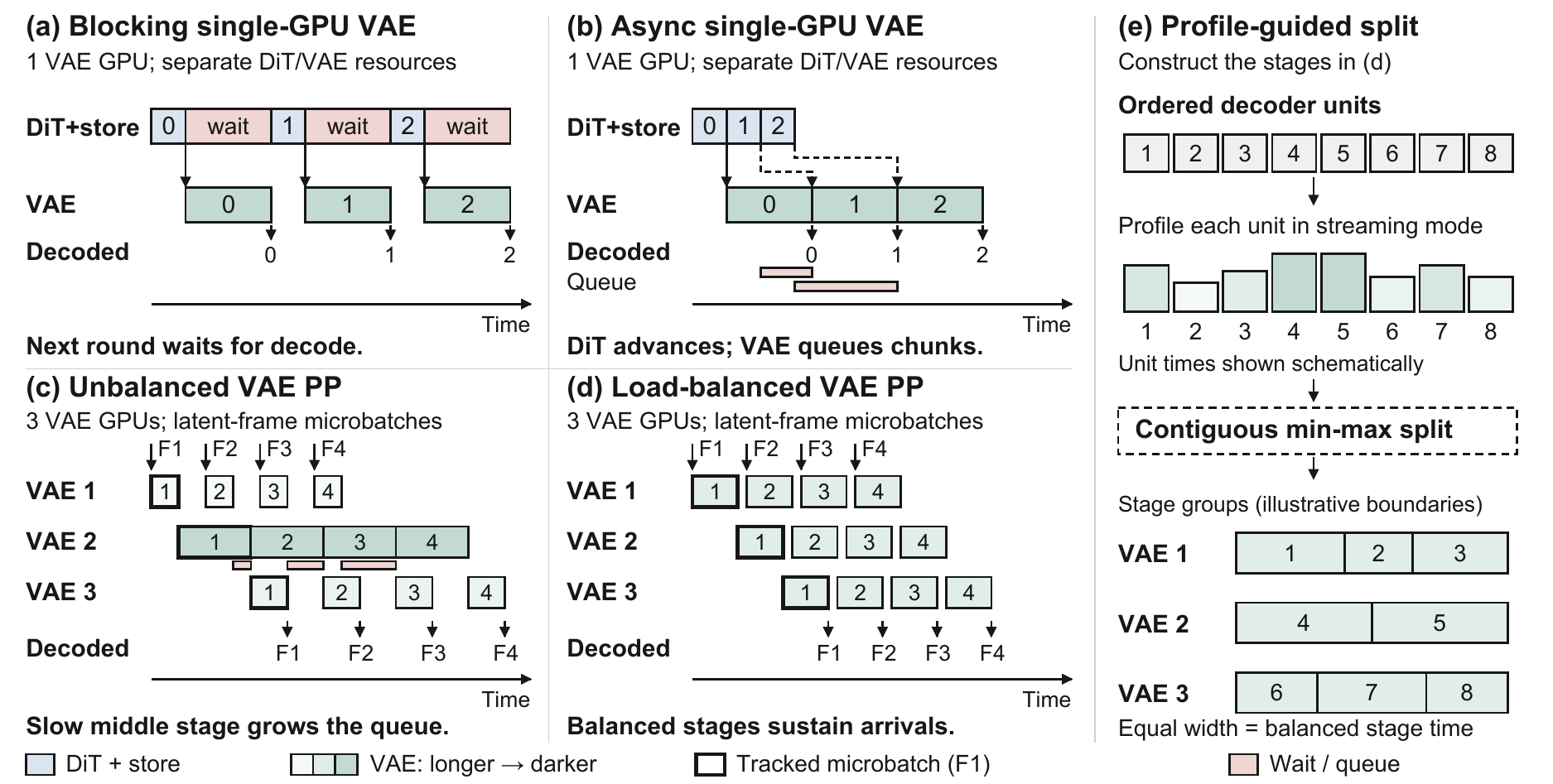}
\caption{Streaming VAE schedules. Panels (a,b) compare blocking and asynchronous decoding with matched costs. Panels (c,d) compare unbalanced and load-balanced PP under identical workloads and arrivals, with darker teal indicating longer stage service times. Panel (e) shows profile-guided contiguous decoder partitions. Each comparison pair uses its own schematic time scale, with transfers omitted and partition boundaries shown schematically. The first output incurs pipeline fill \mbox{latency (Appendix~\ref{sec:streaming-vae})}.}
\label{fig:vae-pipeline}
\end{figure}

Local computation and K/V delivery require joint optimization. SageAttention~\citep{zhang2025sageattention} and W8A8 target faster local operators but alter producer--consumer progress. K/V quantization reduces payloads, and paged writes reduce receiver-side context assembly. The goal is K/V readiness before each layer's attention, reducing exposed communication and \mbox{synchronization waits}.

Asynchronous decoding overlaps VAE and DiT, but the slower pipeline limits E2E throughput. We use profiled streaming times to balance contiguous VAE partitions across dedicated GPUs while preserving request-local causal caches (Fig.~\ref{fig:vae-pipeline}). Decoding and inter-stage transfers must sustain DiT's output rate. The first decoded output still incurs pipeline fill latency (Appendix~\ref{sec:streaming-vae}).

WP uses full-model replicas, with execution width determined by the denoising-step count. Logical K/V fan-out remains quadratic in the number of stages, while attention workloads vary across stages. Appendices~\ref{sec:storage-communication} and~\ref{sec:execution-cost} quantify communication traffic and pipeline fill/drain costs, respectively.

\section{Experiments}
\label{sec:evaluation}

\paragraph{Table notation.}
In all experimental tables, \tabbestinline{teal} marks the best value for each metric within its comparison group, and \tabissueinline{coral} marks unusable outcomes.

\paragraph{Models and training.}
We evaluate 4-step and 5-step Wave Forcing built on Wan2.1-T2V-1.3B~\citep{wan2025wan} against Rolling Forcing~\citep{liu2025rollingforcing} and Causal Forcing~\citep{zhu2026causalforcing}. All methods retain their native denoising schedules and three latent frames per chunk. Training and checkpoint details appear in Sec.~\ref{sec:training} and Appendices~\ref{sec:training-appendix} and~\ref{sec:evaluation-details}.

\paragraph{Evaluation.}
For short videos, we report VBench~\citep{huang2024vbench} Total, Quality, Semantic, and Dynamic Degree, generating one 129-frame video per original prompt. For long videos, we report VBench-Long Quality and Dynamic Degree on 200 MovieGen~\citep{polyak2025moviegen} prompts, generating 100-second videos. All videos use $832\times480$ resolution, 16\,FPS, and seed 42, and benchmark scores follow official normalization and weighting. Tab.~\ref{tab:quality-main} summarizes the results, with the full protocol \mbox{in Appendix~\ref{sec:evaluation-details}}.

\paragraph{Long-video stability diagnostics.}
High motion scores can coexist with subject exit or unidirectional camera drift. We assess these failures on a fixed 100-prompt, subject-focused MovieGen subset shared across methods and training ablations. Subject Preservation measures visibility of the SAM3-tracked~\citep{carion2025sam3} principal entity, with boundary weighting for partial disappearance. Camera Drift penalizes a subject's approach to a frame boundary only with persistent outward background motion in the same direction. Appendix~\ref{sec:evaluation-details} details prompt selection and both metrics.

\paragraph{Compute configurations.}
Execution tests use BF16 with FlashAttention-3 (FA3)~\citep{shah2024flashattention3} as reference. Sage uses SageAttention~\citep{zhang2025sageattention} with BF16 linear layers, while Sage+W8A8 adds 8-bit linear weights and activations. Both Sage tiers use FA3 cross-attention. These labels describe local computation, not K/V transport precision.

\subsection{Generation quality and motion}
\label{sec:quality-results}
\label{sec:quality}
\label{sec:motion-diagnostics}

Tab.~\ref{tab:quality-main} compares short- and long-video generation.

\begin{table}[t]
\centering
\small
\setlength{\tabcolsep}{2.6pt}
\caption{Short- and long-video comparison. Dynamic is Dynamic Degree. Subj./Drift are Subject Preservation/Camera Drift (scores: 0--100). The two FPS columns report DiT-only and end-to-end (E2E) runs.}
\label{tab:paradigm-quality}
\label{tab:quality-main}
\label{tab:long-motion}
\resizebox{\linewidth}{!}{%
\begin{tabular}{@{}l*{10}{c}cc@{}}
\toprule
& & & \multicolumn{4}{c}{\tabhead{Standard VBench $\uparrow$}} & \multicolumn{2}{c}{\tabhead{VBench-Long $\uparrow$}} & \multicolumn{2}{c}{\tabhead{Stability}} & \multicolumn{2}{c}{\tabhead{FPS $\uparrow$}} \\
\cmidrule(lr){4-7}\cmidrule(lr){8-9}\cmidrule(lr){10-11}\cmidrule(lr){12-13}
\tabhead{Method} & \tabhead{Steps} & \tabhead{GPUs} & \tabhead{Total} & \tabhead{Quality} & \tabhead{Semantic} & \tabhead{Dynamic} & \tabhead{Quality} & \tabhead{Dynamic} & \tabhead{Subj.\,$\uparrow$} & \tabhead{Drift\,$\downarrow$} & \tabhead{DiT} & \tabhead{E2E} \\
\midrule
Rolling Forcing & 5 & 1 & 81.15 & 83.77 & 70.66 & 51.39 & 81.18 & 36.63 & 92.75 & \tabbest{0.17} & 29.12 & 17.46 \\
+ Causal Attention & 5 & 8 & 80.89 & 83.35 & \tabbest{71.05} & 37.50 & 80.50 & 46.60 & 91.26 & 0.40 & 104.85 & 96.56 \\
Causal Forcing (Full Attn.) & 4 & 1 & 79.61 & 82.83 & 66.72 & \tabbest{100.00} & \tabbest{82.15} & \tabbest{96.61} & \tabissue{71.11} & \tabissue{11.14} & 33.88 & 19.06 \\
\midrule
\textbf{Wave Forcing (Ours)} & 4 & 8 & 80.74 & 83.93 & 67.95 & 84.72 & 80.89 & 74.97 & 91.88 & 0.36 & 147.58 & \tabbest{134.81} \\
+ SageAttention & & 8 & 81.03 & 84.02 & 69.07 & 65.28 & 81.16 & 77.24 & 90.79 & 0.36 & \tabbest{147.78} & 134.79 \\
\textbf{Wave Forcing (Ours)} & 5 & 8 & \tabbest{81.64} & \tabbest{84.40} & 70.59 & 94.44 & 81.26 & 91.45 & \tabbest{97.10} & 0.37 & 104.82 & 96.46 \\
+ SageAttention & & 8 & 81.02 & 83.92 & 69.41 & 80.56 & 81.14 & 91.14 & 94.43 & 0.37 & 104.80 & 96.48 \\
\bottomrule
\end{tabular}%
}
\end{table}

We compare open-source models of similar scale. Rolling Forcing~\citep{liu2025rollingforcing} is evaluated with its original rolling window and with a causal mask under WaveRT. For Causal Forcing~\citep{zhu2026causalforcing}, we use the full-attention long-video variant with causal ODE initialization from an autoregressive~teacher.

Wave Forcing achieves the best overall balance of quality, motion, and stability (Tab.~\ref{tab:quality-main}). The 5-step model leads VBench Total (81.64) and Quality (84.40), with VBench-Long Quality of 81.26.

\begin{figure}[htbp]
\centering
\includegraphics[width=\linewidth]{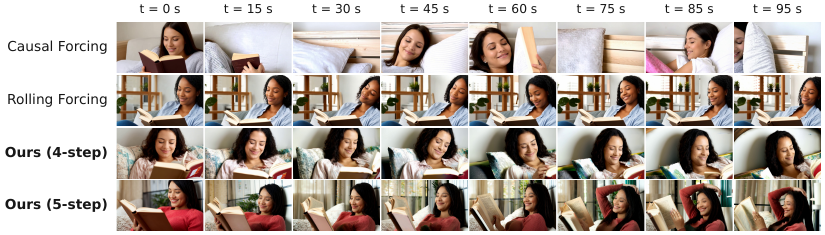}
\caption{Qualitative comparison. Rows show Causal Forcing, Rolling Forcing, and 4-step and 5-step Wave Forcing.}
\label{fig:qualitative-comparison}
\end{figure}

Causal Forcing attains the highest VBench-Long Quality (82.15) and Dynamic Degree (96.61), but its Subject Preservation (71.11) and Camera Drift (11.14) indicate unusable stability. The subject disappears at 30 and 75 seconds, leaving pillows and the headboard in view (Fig.~\ref{fig:qualitative-comparison}). Its high motion score accompanies severe subject loss and camera drift.

\Needspace{3\baselineskip}
Rolling Forcing keeps the subject visible (92.75) with the lowest Camera Drift (0.17), but its VBench-Long Dynamic Degree is only 36.63. Wave Forcing raises Dynamic Degree to 91.45 and Subject Preservation to 97.10 while keeping Camera Drift low at 0.37. Its sampled frames retain the reader while showing changes in pose and interaction with the book (Fig.~\ref{fig:qualitative-comparison}). Wave Forcing combines strong motion, visual quality, and stability without an unusable score on any metric.

\FloatBarrier
\Needspace{17\baselineskip}
\subsection{Training ablations}
\label{sec:training-study}
\label{sec:training-results}

Table~\ref{tab:training-ablation} reports 4-step training ablations for off-policy DMD and each Stage~3 branch.

\begin{wraptable}[11]{r}{0.61\textwidth}
\vspace{-\intextsep}
\centering
\footnotesize
\setlength{\tabcolsep}{2.8pt}
\captionsetup{font=small}
\caption{4-step training ablations on long videos. All rows evaluate final models. Quality/Dynamic are VBench-Long scores.}
\label{tab:training-ablation}
\label{tab:stage3-branch-ablation}
\begin{tabular}{@{}l*{4}{c}@{}}
\toprule
\tabhead{Model} & \tabhead{Quality\,$\uparrow$} & \tabhead{Dynamic\,$\uparrow$} & \tabhead{Subj.\,$\uparrow$} & \tabhead{Drift\,$\downarrow$} \\
\midrule
w/o off-policy DMD & 76.68 & 15.73 & \tabissue{76.53} & 0.26 \\
w/o RF  & 80.19 & \tabbest{88.82} & \tabissue{77.48} & \tabissue{6.24} \\
w/o WF & 80.34 & 48.12 & \tabbest{95.64} & \tabbest{0.10} \\
w/o SF & 75.35 & 23.66 & \tabissue{84.04} & 1.89 \\
\midrule
Ours full        & \tabbest{80.89} & 74.97 & 91.88 & 0.36 \\
\bottomrule
\end{tabular}
\end{wraptable}

\begin{figure}[htbp]
\centering
\includegraphics[width=\linewidth]{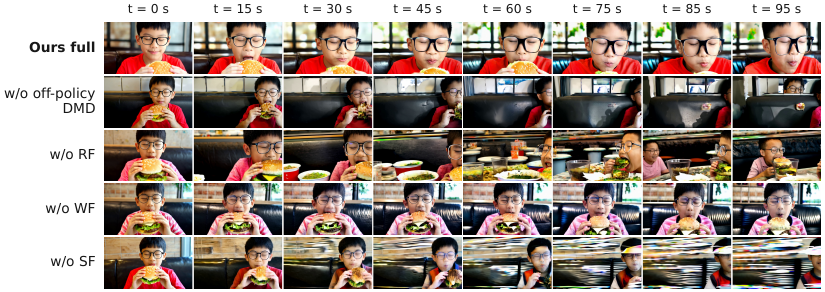}
\caption{Qualitative training ablations. We compare the full 4-step model with each training ablation.}
\label{fig:qualitative-ablation}
\end{figure}

\paragraph{Off-policy DMD.}
Skipping Stage~2 and proceeding directly to on-policy adaptation in Stage~3 reduces Quality from 80.89 to 76.68 and Dynamic Degree from 74.97 to 15.73. Subject Preservation also falls to 76.53, and the principal subject moves out of view (Fig.~\ref{fig:qualitative-ablation}). Off-policy DMD prepares the model to preserve both visual quality and motion during causal adaptation.

\paragraph{Mixed Training Strategy.}
Removing SF reduces Quality to 75.35 and Dynamic Degree to 23.66, with visible degradation and subject loss (Fig.~\ref{fig:qualitative-ablation}). Removing RF increases Dynamic Degree to 88.82, but Subject Preservation falls to 77.48 and Camera Drift rises to 6.24, the higher motion score reflects unstable framing and subject loss. Removing WF yields the highest Subject Preservation (95.64) and lowest Camera Drift (0.10), but lowers Dynamic Degree to 48.12.

The full model provides the best overall balance, combining the highest Quality (80.89) with strong Dynamic Degree (74.97), high Subject Preservation (91.88), and low Camera Drift (0.36). It sustains motion and subject stability without an unusable score on any metric. We therefore retain all three branches in the final training recipe.

\Needspace{8\baselineskip}
\subsection{End-to-end inference efficiency}
\label{sec:inference-results}

We measure 4-step WaveRT at the resolution of $832\times480$ on a single $8\times$H200 141-GB node, allocating four GPUs to denoising, one to store, and three to VAE decoding. These runs use randomly initialized (dummy) DiT weights, since the corresponding pretrained weights are unavailable and we lack sufficient compute resources to adequately train models at the 14B scale. We report FPS as 1593 decoded frames divided by the median elapsed time over three runs and compute tick p50 from pooled steady-state ticks. DiT timing covers diffusion, communication, store, and pipeline fill/drain, while E2E timing extends through VAE completion. Because VAE decoding overlaps DiT execution, the per-run gap between these timings measures the exposed decoding tail. Loading and file encoding are \mbox{excluded (Appendix~\ref{sec:results-appendix})}.

\begin{figure}[t]
\centering
\includegraphics[width=\linewidth]{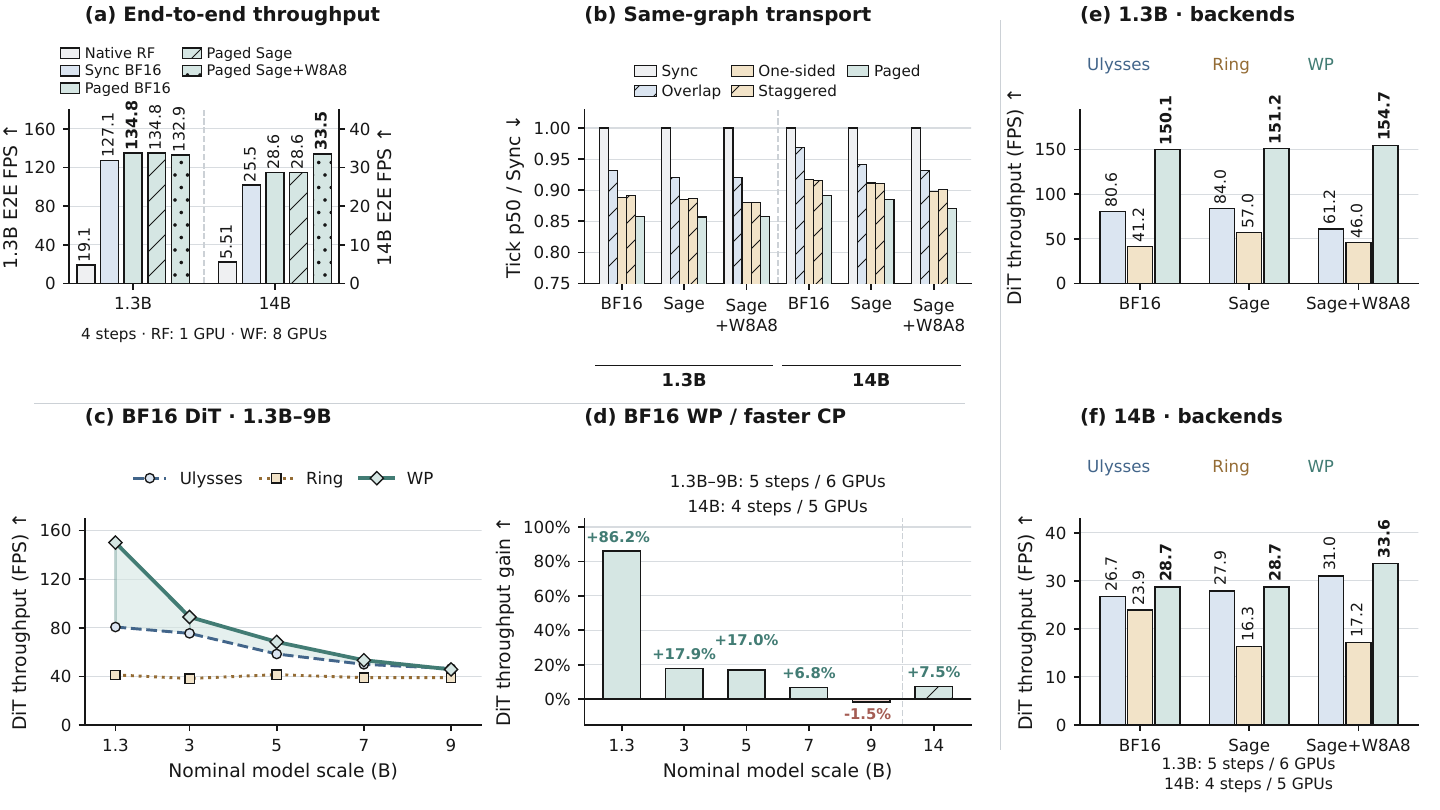}
\caption{System efficiency with dummy DiT weights. (a) E2E throughput of 4-step WaveRT (8 GPUs) and the single-GPU baseline, with separate axes for 1.3B and 14B. (b) Steady-state tick p50 on the same WF graph, normalized to Sync. (c,d) BF16 DiT throughput across scales and WP's percentage gain over the faster CP baseline. (e,f) DiT throughput across compute backends at 1.3B and 14B. Panels (c--f) disable VAE and use 5 steps/6 GPUs for 1.3B--9B and 4 steps/5 GPUs for 14B. Full measurements and repetition details appear in Tabs.~\ref{tab:runtime-core} and~\ref{tab:cp-dit}.}
\label{fig:systems-performance}
\end{figure}

Fig.~\ref{fig:systems-performance}a compares runtime and compute configurations. At 1.3B, WaveRT improves BF16 E2E throughput from 127.06 FPS with Sync to 134.81 FPS (6.1\%) on the same WF graph. WaveRT + Sage reaches 134.79 E2E FPS versus 147.78 DiT FPS in the same runs (Tab.~\ref{tab:runtime-core}), essentially matching BF16 E2E throughput. Adding W8A8 lowers tick p50 but reduces E2E throughput to 132.89 FPS.
At 14B, the corresponding E2E rates are 25.47, 28.64, 28.62, and 33.50 FPS, versus a 5.51-FPS single-GPU reference. Sage alone is again nearly unchanged, whereas Sage+W8A8 is 17.0\% faster than BF16. The best single-GPU-relative gains ($7.07\times$ at 1.3B and $6.08\times$ at 14B) combine changes to parallelism, the graph, kernels, and decoding.


\subsection{System ablations}
\label{sec:system-ablations}

Fig.~\ref{fig:systems-performance}b fixes the model, WF graph, schedule, topology, and local compute per group, normalizing tick p50 to Sync. Sync gathers layer K/V before attention, whereas Overlap publishes it early. One-sided uses CUDA IPC P2P, Staggered prefetches readiness checks, and Paged writes directly into final \mbox{attention-buffer slots}.

Paged has the lowest tick p50 in all six groups. Sync-to-Paged lowers BF16 tick p50 from 91.96 to 78.88\,ms (14.2\%) at 1.3B and from 464.27 to 413.80\,ms (10.9\%) at 14B. The largest adjacent reduction depends on the group, while at 14B BF16 it is Overlap-to-One-sided. Appendix~\ref{sec:results-appendix} gives repeat ranges and diagnostics. VAE profiles give 73.15/60.91/80.89\,ms, excluding transfers and queueing (Tab.~\ref{tab:mechanism-diagnostics}).
\subsection{Comparison with traditional context parallelism}
\label{sec:cp-comparison}

Fig.~\ref{fig:systems-performance}c,d compares BF16 DiT throughput for attention-level CP (Ulysses and Ring)~\citep{jacobs2023deepspeedulysses,liu2023ringattention} and WP's chunk--stage parallelism across six scales. All methods use 5 steps/6 GPUs for 1.3B--9B and 4 steps/5 GPUs for 14B, satisfying Ulysses' head-divisibility requirement. WP assigns one GPU per denoising stage plus store. Using the architectures listed in Tab.~\ref{tab:synthetic-arch}, we measure throughput with dummy weights and VAE disabled. We report the mean per-run FPS, including communication and synchronization. Repetition details are given in Appendix~\ref{sec:parallelism-analysis}.


In the BF16 results, WP leads at five of six scales, with WP/CP ratios of $0.985$--$1.862\times$. At 1.3B, WP reaches 150.09 FPS versus 80.60 for Ulysses and 41.19 for Ring. At 9B, WP's 45.70 FPS trails Ulysses' 46.41 by 1.53\%. At 14B, WP reaches 28.71 versus 26.72 FPS for the faster CP. Different step counts, GPU budgets, and VAE concurrency preclude attributing cross-panel FPS differences solely to decoding.

Fig.~\ref{fig:systems-performance}e,f compares backends at 1.3B and 14B, with results for all six scales reported in Tab.~\ref{tab:cp-dit}. The three-run Sage+W8A8 measurements put WP first at every scale ($1.025$--$2.530\times$ the faster CP). These results show that WP's Sage-only advantage varies across scales.

\section{Related Work}
\label{sec:related}

\paragraph{Efficient streaming video generation.}
Self Forcing~\citep{huang2025selfforcing} reduces exposure bias by training autoregressive video diffusion on self-generated histories. SGF~\citep{zhuang2026sgf} restores gradient supervision for historical K/V through two-pass replay, while SGF+~\citep{su2026sgfplus} separates context-writing and denoising parameters to reduce gradient conflict. Other approaches include few-step causal distillation~\citep{yin2024causvid,zhu2026causalforcing} and mixed-noise denoising~\citep{liu2025rollingforcing}. AnyFlow~\citep{gu2026anyflow} learns two-time flow maps with on-policy distillation for variable sampling budgets; Mask Forcing~\citep{zhao2026maskforcing} perturbs DMD self-rollouts with dual-noise masking to improve teacher-mode coverage. For long-horizon streaming, Helios~\citep{yuan2026helios} compresses historical and noisy context and simulates drifting during training. Vidu S1~\citep{zhang2026vidus1} uses stage-aware noisy and clean history caches for real-time, voice-controlled character generation. Vidu S2~\citep{zhang2026vidus2} introduces Self-Replay Forcing and supports dynamic references and streaming editing. These methods motivate studying training objectives together with the context available during streaming inference.

\paragraph{Multi-chunk generation.}
Blockwise diffusion also appears in language modeling and action generation. D2F~\citep{wang2025d2f} conditions language generation on partially denoised predecessors, while MBD-LMs~\citep{jin2026mbdlms} align training with concurrent decoding of a bounded set of blocks at heterogeneous noise levels. In vision-language-action models, BlockVLA~\citep{wang2026blockvla} adapts pretrained autoregressive models through block diffusion finetuning, and TBD-VLA~\citep{lee2026tbdvla} organizes actions into temporal blocks and supports asynchronous chunk execution through temporal inpainting. Multimodal Diffusion Forcing~\citep{huang2025mdf} assigns noise levels across time and modalities to model dependencies within multimodal robot trajectories. These approaches illustrate different choices of block dependencies; WF specifically trains video chunks to consume cleaner intermediate context through a directed causal wavefront.

\paragraph{Parallel execution.}
Ulysses~\citep{jacobs2023deepspeedulysses} and Ring~\citep{liu2023ringattention} partition individual forward passes, using all-to-all exchange and K/V circulation, respectively. Live Avatar and HiAR~\citep{huang2025liveavatar,zou2026hiar} explore pipelined execution with fixed-timestep or matched-noise designs. WP instead schedules chunk--stage tasks on full-model replicas (Sec.~\ref{sec:cp-comparison}), adapting wavefront scheduling~\citep{hou2018wavefront,jin2016hybridwavefront} alongside optimized attention and runtimes~\citep{dao2022flashattention,dao2023flashattention2,zhang2025sageattention,zheng2024sglang}.

\paragraph{Video evaluation.}
VBench~\citep{huang2024vbench} provides standardized quality, semantic, and motion measures. Our long-video comparisons (Sec.~\ref{sec:quality-results}) show that high Quality and Dynamic Degree can coexist with subject loss and camera drift. We therefore supplement benchmark scores with targeted stability diagnostics; their scope and future alternatives are discussed in Secs.~\ref{sec:discussion} and~\ref{sec:future-work}.

These lines frame WF's joint design of causal conditioning, training, and execution, together with the evidence used to assess it. Section~\ref{sec:discussion} examines the remaining methodological and system limits; Sec.~\ref{sec:future-work} connects them to better evaluation, interactive generation, and world modeling.

\begingroup
\setlength{\parskip}{3.5pt}
\section{Limitations}
\label{sec:discussion}

Our evaluation establishes streaming generation under fixed conditioning and the tested model, sampling, and hardware configurations. The following limits of the training and execution design discussed in Sec.~\ref{sec:related} motivate the agenda in Sec.~\ref{sec:future-work}.

\paragraph{Quality evaluation and diagnostic coverage.}
VBench-based scores do not fully characterize long-stream quality: in our comparisons, high Quality and Dynamic Degree coexist with subject disappearance and camera drift (Sec.~\ref{sec:quality-results}). Subject Preservation and Camera Drift expose these failures, but rely on a 100-prompt subject-focused subset, tracking and identity association, and geometric heuristics (Appendix~\ref{sec:diagnostic-implementation}). Their sensitivity to thresholds and missing observations, coverage of intentional scene changes, and agreement with human judgments remain to be established. They provide targeted diagnostics, with broader perceptual and semantic quality still requiring assessment.

\paragraph{Output-path capacity and attribution.}
\label{sec:vae-limitations}
VAE pipelining overlaps and partitions decoding but does not remove its finite service capacity. The 1.3B BF16 configurations reach 147.58/134.81 DiT/E2E FPS at four steps and 104.82/96.46 at five steps (Tab.~\ref{tab:quality-main}), with three and two VAE GPUs under the eight-GPU budget, respectively. These gaps are consistent with residual decoder/output-path limits. Aggregate timings, however, do not isolate VAE compute, transfers, queueing, or backpressure, and system DiT timing may reflect decoder interference. For 14B BF16, the gap narrows to 28.74 versus 28.64 DiT/E2E FPS (Tab.~\ref{tab:runtime-core}). Matched VAE-on/off runs and stage-level measurements are needed to identify bottlenecks and guide end-to-end resource allocation.

\paragraph{Training complexity.}
WF currently uses a complex multi-stage training procedure. An important direction is to simplify this procedure while preserving generation quality and long-video stability.

\paragraph{Scalability at a fixed sampling budget.}
For $S$ DiT denoising steps, WP uses $S$ denoising stages and one cache-refresh stage, each with a full-model replica. Stage parallelism is therefore bounded at a fixed step count, limiting the scale of a single pipeline. Our experiments also show that WP's throughput advantage over CP narrows for larger models. These constraints motivate studying finer-grained or hybrid parallelism and resource sharing across streams.

\paragraph{Interaction and serving scope.}
High steady-state throughput alone does not establish low input-to-visible-output latency. Online condition changes require deciding when inputs take effect in active chunks and how cached state is refreshed; the fixed-context equivalence argument in Appendix~\ref{sec:semantic-correctness} does not cover these updates. Prompt switching, camera/action control, closed-loop tasks, and multi-stream deadline guarantees require dedicated evaluation beyond the present fixed-conditioning experiments.

\section{Future Work}
\label{sec:future-work}

The limits in Sec.~\ref{sec:discussion} define our research agenda: strengthen end-to-end streaming execution, enable online interaction, and then extend toward interactive world models. Serving research and richer quality evaluation support these capability extensions throughout. Appendix~\ref{sec:broader-connections} develops the task-specific questions.

\paragraph{End-to-end execution and scalability.}
We plan to profile decoder stages, transfers, queues, and backpressure to guide DiT/VAE resource allocation. To address WP's fixed-step scaling limit, we will study finer-grained execution and combinations with context parallelism, checking communication costs and causal dependencies under each design. Simplifying WF's training recipe while preserving motion and long-horizon stability remains a parallel objective.

\paragraph{Serving concurrent streams.}
StreamDiffusion's image-stream batching~\citep{kodaira2025streamdiffusion} and StreamDiffusionV2's video-stream scheduling~\citep{feng2026streamdiffusionv2} motivate latency-aware batching. SlackServe~\citep{xia2026slackserve} uses playout slack for resource and quality decisions, while PhysisServe~\citep{physis2026physisserve} explores fine-grained stage pipelining. Building on these directions, we aim to study admission control, batching, and resource allocation for concurrent WF streams while respecting their context dependencies. Evaluation should include startup delay, output jitter, playback stalls, fairness, and quality alongside throughput.

\paragraph{Interactive streaming generation.}
The next capability is updating prompts during generation. LongLive's KV recaching~\citep{yang2025longlive} provides a starting point for coordinating new conditions with visual history. Extending WF requires defining which active chunks receive each update, refreshing affected state, and adapting training to condition changes. We will assess prompt-to-visible-output latency, prompt adherence, and visual continuity, rather than inferring responsiveness from output FPS.

\paragraph{Interactive world models.}
Camera and action inputs further introduce feedback-dependent state evolution. We aim to explore both learned world prediction~\citep{zhao2026minwm,su2026onewm} and systems in which code or an engine maintains world state while a video model renders observations~\citep{chen2026codeworldmodel,zhang2026matrixgame4}. The latter motivates adapting WF for structure-conditioned streaming rendering. Both routes require training under the corresponding control interface and evaluating control alignment, state consistency, and closed-loop behavior. They build on the execution and interaction work above and remain prospective extensions of WF.

\paragraph{Richer quality evaluation.}
To address the metric limits above, we plan to incorporate streaming-interaction benchmarks such as StreamEval~\citep{streameval2026}, which evaluates task execution and state consistency under timed inputs alongside visual quality. Long-video understanding methods such as StreamMeCo~\citep{wang2026streammeco} could help organize temporal evidence for an auxiliary quality judge. We aim to combine automatic diagnostics, understanding-based judgments, and human assessment, with explicit criteria for subject continuity, intended motion, and instruction execution. Such judges require timestamped evidence, calibration against human annotations, and checks for missed failures before their scores can support model comparisons.

\Needspace{13\baselineskip}
\section{Conclusion}
\label{sec:conclusion}

We present Wave Forcing, which jointly designs causal context, training strategies, and multi-GPU pipelining to denoise multiple video chunks concurrently. WF uses cleaner intermediate states of preceding chunks as context for subsequent chunks, with three-stage training adapting the model to mixed-noise contexts and self-generated histories. A pipeline of full-model replicas then produces one video chunk per concurrent forward round after fill, while WaveRT overlaps K/V delivery with computation to reduce dependency waits. Experiments show that WF improves motion while maintaining overall generation quality and achieves high-throughput streaming video generation. These results demonstrate that jointly designing generation context, training, and execution can translate cross-chunk parallelism into practical gains in video generation efficiency.

\label{page:main-end}
\FloatBarrier
\Needspace{16\baselineskip}

\section*{Author Contributions}
\label{sec:author-contributions}

\begingroup
\raggedright
\setlength{\parskip}{3.5pt}
\newcommand{\contribution}[2]{\noindent\textbf{#1:} #2\par}

\contribution{Core idea}{%
\mbox{Yijie~Jin}}

\contribution{Feasibility exploration}{%
\mbox{Yijie~Jin}\textsuperscript{*}, \mbox{Jiajun~Xu}\textsuperscript{*}}

\contribution{Training methods \& infrastructure optimization}{%
\mbox{Yuxuan~Liu}, \mbox{Yijie~Jin}, \mbox{Ye~Huang}}

\contribution{Inference implementation \& infrastructure optimization}{%
\mbox{Yijie~Jin}\textsuperscript{*}, \mbox{Jiajun~Xu}\textsuperscript{*}, \mbox{\mbox{Jiayang~Li}, \mbox{Chenling~Hong}}}

\contribution{Experiments}{%
\mbox{Yuxuan~Liu}\textsuperscript{*}, \mbox{Jiajun~Xu}\textsuperscript{*}, \mbox{Yijie~Jin}, \mbox{Chuzhi~Wang}, \mbox{Ye~Huang}}

\contribution{Paper Writing}{
\mbox{Yijie~Jin}\textsuperscript{*}, \mbox{Yuxuan~Liu}\textsuperscript{*}, \mbox{Jiajun~Xu}, \mbox{Ruiyang~Chen}
}

\contribution{Technical consultation}{%
\mbox{Xu~Wang}, \mbox{Ye~Huang}}

\contribution{Project leadership}{%
\mbox{Zhijie~Deng}}

\vspace{1pt}
{\footnotesize\textit{\textsuperscript{*} Equal contribution within the corresponding category.}\par}
\endgroup

\section*{AI Use Statement}
\label{sec:ai-use}
AI assistants were used to support conceptual framing and manuscript organization, discuss methodology and systems analogies, interpret available implementation and experimental evidence, translate and edit text, and develop and revise figures. This assistance does not constitute independent verification of the reported results. The authors remain responsible for the paper's claims, citations, and disclosures.
\par

\endgroup

\bibliographystyle{plainnat}
\bibliography{reference}

@article{wang2025d2f,
  title={Diffusion {LLMs} Can Do Faster-Than-{AR} Inference via Discrete Diffusion Forcing},
  author={Wang, Xu and Xu, Chenkai and Jin, Yijie and Jin, Jiachun and Zhang, Hao and Deng, Zhijie},
  journal={arXiv preprint arXiv:2508.09192},
  year={2025},
  url={https://arxiv.org/abs/2508.09192}
}

@article{jin2026mbdlms,
  title={Multi-Block Diffusion Language Models},
  author={Jin, Yijie and Xu, Jiajun and Liu, Yuxuan and Xu, Chenkai and Tu, Yi and Li, Jiajun and Tu, Dandan and Yan, Xiaohui and Yu, Kai and Liu, Pengfei and Deng, Zhijie},
  journal={arXiv preprint arXiv:2606.29215},
  year={2026},
  url={https://arxiv.org/abs/2606.29215}
}

@article{wang2026blockvla,
  title={{BlockVLA}: Accelerating Autoregressive {VLA} via Block Diffusion Finetuning},
  author={Wang, Ruiheng and Bai, Shuanghao and Zhang, Haoran and Chen, Badong and Xu, Xiangyu},
  journal={arXiv preprint arXiv:2605.13382},
  year={2026},
  url={https://arxiv.org/abs/2605.13382}
}

@article{lee2026tbdvla,
  title={{TBD-VLA}: Temporal Block Diffusion Vision Language Action Model},
  author={Lee, Sung-Wook and Kang, Xuhui and Kuo, Yen-Ling},
  journal={arXiv preprint arXiv:2606.07895},
  year={2026},
  url={https://arxiv.org/abs/2606.07895}
}

@article{huang2025mdf,
  title={Multimodal Diffusion Forcing for Forceful Manipulation},
  author={Huang, Zixuan and Hou, Huaidian and Berenson, Dmitry},
  journal={arXiv preprint arXiv:2511.04812},
  year={2025},
  url={https://arxiv.org/abs/2511.04812}
}

@article{zhuang2026sgf,
  title={Self Gradient Forcing: Native Long Video Extrapolation},
  author={Zhuang, Junhao and Zhang, Shiyi and Bian, Yuxuan and Li, Yaowei and Luo, Yawen and Liu, Yijun and Jin, Weiyang and Zhang, Songchun and He, Xianglong and Zhang, Xuying and Li, Haoran and Huang, Haoyang and Xue, Zeyue and Duan, Nan},
  journal={arXiv preprint arXiv:2607.20368},
  year={2026},
  url={https://arxiv.org/abs/2607.20368}
}

@article{su2026sgfplus,
  title={{SGF+}: Decoupling Gradient Flows for Autoregressive Video Generation},
  author={Su, Zihan and Zhuang, Junhao and Li, Yaowei and Lu, Siwen and Li, Haoran and Li, Lingen and Wu, Haoyu and Jin, Weiyang and Zhang, Songchun and Huang, Haoyang and Yuan, Chun and Xue, Zeyue and Duan, Nan},
  journal={arXiv preprint arXiv:2610.10429},
  year={2026},
  url={https://arxiv.org/abs/2610.10429}
}

@article{yuan2026helios,
  title={{Helios}: Real Real-Time Long Video Generation Model},
  author={Yuan, Shenghai and Yin, Yuanyang and Li, Zongjian and Huang, Xinwei and Yang, Xiao and Yuan, Li},
  journal={arXiv preprint arXiv:2603.04379},
  year={2026},
  url={https://arxiv.org/abs/2603.04379}
}

@article{zeng2026lpm,
  title={{LPM} 1.0: Video-based Character Performance Model},
  author={Zeng, Ailing and Yang, Casper and Ge, Chauncey and Zhang, Eddie and Xu, Garvey and Lin, Gavin and Gu, Gilbert and Pi, Jeremy and Li, Leo and Shi, Mingyi and Wang, Shawn and Bi, Sheng and Tang, Steven and Hang, Thorn and Guo, Tobey and Li, Vincent and Tong, Xin and Li, Yikang and Sun, Yuchen and Zhao, Yue and Lu, Yuhan and Li, Yuwei and Zhang, Zane and Yang, Zeshi and Ye, Zi},
  journal={arXiv preprint arXiv:2604.07823},
  year={2026},
  url={https://arxiv.org/abs/2604.07823}
}

@inproceedings{zhang2018lpips,
  title={The Unreasonable Effectiveness of Deep Features as a Perceptual Metric},
  author={Zhang, Richard and Isola, Phillip and Efros, Alexei A. and Shechtman, Eli and Wang, Oliver},
  booktitle={Proceedings of the IEEE Conference on Computer Vision and Pattern Recognition},
  pages={586--595},
  year={2018},
  organization={IEEE},
  url={https://openaccess.thecvf.com/content_cvpr_2018/html/Zhang_The_Unreasonable_Effectiveness_CVPR_2018_paper.html}
}

@inproceedings{huang2024vbench,
  title={{VBench}: Comprehensive Benchmark Suite for Video Generative Models},
  author={Huang, Ziqi and He, Yinan and Yu, Jiashuo and Zhang, Fan and Si, Chenyang and Jiang, Yuming and Zhang, Yuanhan and Wu, Tianxing and Jin, Qingyang and Chanpaisit, Nattapol and Wang, Yaohui and Chen, Xinyuan and Wang, Limin and Lin, Dahua and Qiao, Yu and Liu, Ziwei},
  booktitle={Proceedings of the IEEE/CVF Conference on Computer Vision and Pattern Recognition},
  pages={21807--21818},
  year={2024},
  organization={IEEE},
  url={https://github.com/Vchitect/VBench}
}

@article{huang2025vbenchplus,
  title={{VBench++}: Comprehensive and Versatile Benchmark Suite for Video Generative Models},
  author={Huang, Ziqi and Zhang, Fan and Xu, Xiaojie and He, Yinan and Yu, Jiashuo and Dong, Ziyue and Ma, Qianli and Chanpaisit, Nattapol and Si, Chenyang and Jiang, Yuming and Wang, Yaohui and Chen, Xinyuan and Chen, Ying-Cong and Wang, Limin and Lin, Dahua and Qiao, Yu and Liu, Ziwei},
  journal={IEEE Transactions on Pattern Analysis and Machine Intelligence},
  year={2026},
  volume={48},
  number={3},
  pages={3268-3285},
  doi={10.1109/TPAMI.2025.3633890}
}

@inproceedings{yin2024dmd,
  title={One-step Diffusion with Distribution Matching Distillation},
  author={Yin, Tianwei and Gharbi, Micha{\"e}l and Zhang, Richard and Shechtman, Eli and Durand, Fr{\'e}do and Freeman, William T. and Park, Taesung},
  booktitle={Proceedings of the IEEE/CVF Conference on Computer Vision and Pattern Recognition},
  pages={6613--6623},
  year={2024},
  organization={IEEE},
  url={https://openaccess.thecvf.com/content/CVPR2024/html/Yin_One-step_Diffusion_with_Distribution_Matching_Distillation_CVPR_2024_paper.html}
}

@inproceedings{yin2024causvid,
  title={From Slow Bidirectional to Fast Autoregressive Video Diffusion Models},
  author={Yin, Tianwei and Zhang, Qiang and Zhang, Richard and Freeman, William T. and Durand, Fredo and Shechtman, Eli and Huang, Xun},
  booktitle={Proceedings of the IEEE/CVF Conference on Computer Vision and Pattern Recognition},
  pages={22963--22974},
  year={2025},
  organization={IEEE},
  url={https://openaccess.thecvf.com/content/CVPR2025/html/Yin_From_Slow_Bidirectional_to_Fast_Autoregressive_Video_Diffusion_Models_CVPR_2025_paper.html}
}

@inproceedings{huang2025selfforcing,
  title={Self Forcing: Bridging the Train-Test Gap in Autoregressive Video Diffusion},
  author={Huang, Xun and Li, Zhengqi and He, Guande and Zhou, Mingyuan and Shechtman, Eli},
  booktitle={Advances in Neural Information Processing Systems},
  volume={38},
  pages={167283--167308},
  year={2025},
  doi={10.52202/085713-5576},
  url={https://proceedings.neurips.cc/paper_files/paper/2025/hash/f4823f831af67a3ef15e41a85434422a-Abstract-Conference.html}
}

@inproceedings{liu2025rollingforcing,
  title={Rolling Forcing: Autoregressive Long Video Diffusion in Real Time},
  author={Liu, Kunhao and Hu, Wenbo and Xu, Jiale and Shan, Ying and Lu, Shijian},
  booktitle={International Conference on Learning Representations},
  pages={91177--91196},
  year={2026},
  url={https://proceedings.iclr.cc/paper_files/paper/2026/hash/935151cc6cb5d8b6816133b75233775a-Abstract-Conference.html},
}

@article{zhu2026causalforcing,
  title={Causal Forcing: Autoregressive Diffusion Distillation Done Right for High-Quality Real-Time Interactive Video Generation},
  author={Zhu, Hongzhou and Zhao, Min and He, Guande and Su, Hang and Li, Chongxuan and Zhu, Jun},
  journal={arXiv preprint arXiv:2602.02214},
  year={2026},
  url={https://arxiv.org/abs/2602.02214}
}

@inproceedings{dao2022flashattention,
  title={{FlashAttention}: Fast and Memory-Efficient Exact Attention with {IO-Awareness}},
  author={Dao, Tri and Fu, Daniel Y. and Ermon, Stefano and Rudra, Atri and R{\'e}, Christopher},
  booktitle={Advances in Neural Information Processing Systems},
  volume={35},
  pages={16344--16359},
  year={2022},
  doi={10.52202/068431-1189},
  url={https://proceedings.neurips.cc/paper_files/paper/2022/hash/67d57c32e20fd0a7a302cb81d36e40d5-Abstract.html}
}

@inproceedings{dao2023flashattention2,
  title={{FlashAttention-2}: Faster Attention with Better Parallelism and Work Partitioning},
  author={Dao, Tri},
  booktitle={International Conference on Learning Representations},
  pages={35549--35562},
  year={2024},
  url={https://proceedings.iclr.cc/paper_files/paper/2024/hash/98ed250b203d1ac6b24bbcf263e3d4a7-Abstract-Conference.html}
}

@inproceedings{zhang2025sageattention,
  title={{SageAttention}: Accurate 8-Bit Attention for Plug-and-Play Inference Acceleration},
  author={Zhang, Jintao and Wei, Jia and Zhang, Pengle and Zhu, Jun and Chen, Jianfei},
  booktitle={International Conference on Learning Representations},
  pages={71566--71585},
  year={2025},
  url={https://proceedings.iclr.cc/paper_files/paper/2025/hash/b286c344d38e10d2466c0514b78e2f36-Abstract-Conference.html}
}

@inproceedings{zheng2024sglang,
  title={{SGLang}: Efficient Execution of Structured Language Model Programs},
  author={Zheng, Lianmin and Yin, Liangsheng and Xie, Zhiqiang and Sun, Chuyue and Huang, Jeff and Yu, Cody Hao and Cao, Shiyi and Kozyrakis, Christos and Stoica, Ion and Gonzalez, Joseph E. and Barrett, Clark and Sheng, Ying},
  booktitle={Advances in Neural Information Processing Systems},
  volume={37},
  pages={62557--62583},
  year={2024},
  doi={10.52202/079017-2000},
  url={https://proceedings.neurips.cc/paper_files/paper/2024/hash/724be4472168f31ba1c9ac630f15dec8-Abstract-Conference.html}
}

@inproceedings{hou2018wavefront,
  title={Highly Efficient Compensation-Based Parallelism for Wavefront Loops on {GPUs}},
  author={Hou, Kaixi and Wang, Hao and Feng, Wu-chun and Vetter, Jeffrey S. and Lee, Seyong},
  booktitle={IEEE International Parallel and Distributed Processing Symposium},
  pages={276--285},
  year={2018},
  organization={IEEE}
}

@article{jin2016hybridwavefront,
  title={Hybrid Static/Dynamic Schedules for Tiled Polyhedral Programs},
  author={Jin, Tian and Prajapati, Nirmal and Ranasinghe, Waruna and Iooss, Guillaume and Zou, Yun and Rajopadhye, Sanjay and Wonnacott, David},
  journal={arXiv preprint arXiv:1610.07236},
  year={2016},
  url={https://arxiv.org/abs/1610.07236}
}

@article{wan2025wan,
  title={Wan: Open and Advanced Large-Scale Video Generative Models},
  author={Wan, Team and Wang, Ang and Ai, Baole and Wen, Bin and Mao, Chaojie and Xie, Chen-Wei and Chen, Di and Yu, Feiwu and Zhao, Haiming and Yang, Jianxiao and Zeng, Jianyuan and Wang, Jiayu and Zhang, Jingfeng and Zhou, Jingren and Wang, Jinkai and Chen, Jixuan and Zhu, Kai and Zhao, Kang and Yan, Keyu and Huang, Lianghua and Feng, Mengyang and Zhang, Ningyi and Li, Pandeng and Wu, Pingyu and Chu, Ruihang and Feng, Ruili and Zhang, Shiwei and Sun, Siyang and Fang, Tao and Wang, Tianxing and Gui, Tianyi and Weng, Tingyu and Shen, Tong and Lin, Wei and Wang, Wei and Wang, Wei and Zhou, Wenmeng and Wang, Wente and Shen, Wenting and Yu, Wenyuan and Shi, Xianzhong and Huang, Xiaoming and Xu, Xin and Kou, Yan and Lv, Yangyu and Li, Yifei and Liu, Yijing and Wang, Yiming and Zhang, Yingya and Huang, Yitong and Li, Yong and Wu, You and Liu, Yu and Pan, Yulin and Zheng, Yun and Hong, Yuntao and Shi, Yupeng and Feng, Yutong and Jiang, Zeyinzi and Han, Zhen and Wu, Zhi-Fan and Liu, Ziyu},
  journal={arXiv preprint arXiv:2503.20314},
  year={2025},
  url={https://arxiv.org/abs/2503.20314}
}

@inproceedings{yang2025longlive,
  title={{LongLive}: Real-time Interactive Long Video Generation},
  author={Yang, Shuai and Huang, Wei and Chu, Ruihang and Xiao, Yicheng and Zhao, Yuyang and Wang, Xianbang and Li, Muyang and Xie, Enze and Chen, Yingcong and Lu, Yao and Han, Song and Chen, Yukang},
  booktitle={International Conference on Learning Representations},
  pages={90037--90059},
  year={2026},
  url={https://proceedings.iclr.cc/paper_files/paper/2026/hash/91a1610c6ed9e02d33f826b46f472b92-Abstract-Conference.html},
}

@article{jacobs2023deepspeedulysses,
  title={{DeepSpeed Ulysses}: System Optimizations for Enabling Training of Extreme Long Sequence Transformer Models},
  author={Jacobs, Sam Ade and Tanaka, Masahiro and Zhang, Chengming and Zhang, Minjia and Song, Shuaiwen Leon and Rajbhandari, Samyam and He, Yuxiong},
  journal={arXiv preprint arXiv:2309.14509},
  year={2023},
  url={https://arxiv.org/abs/2309.14509}
}

@inproceedings{shah2024flashattention3,
  title={{FlashAttention-3}: Fast and Accurate Attention with Asynchrony and Low-precision},
  author={Shah, Jay and Bikshandi, Ganesh and Zhang, Ying and Thakkar, Vijay and Ramani, Pradeep and Dao, Tri},
  booktitle={Advances in Neural Information Processing Systems},
  volume={37},
  pages={68658--68685},
  year={2024},
  doi={10.52202/079017-2193},
  url={https://proceedings.neurips.cc/paper_files/paper/2024/hash/7ede97c3e082c6df10a8d6103a2eebd2-Abstract-Conference.html}
}

@inproceedings{liu2023ringattention,
  title={{RingAttention} with Blockwise Transformers for Near-Infinite Context},
  author={Liu, Hao and Zaharia, Matei and Abbeel, Pieter},
  booktitle={International Conference on Learning Representations},
  pages={3992--4008},
  year={2024},
  url={https://proceedings.iclr.cc/paper_files/paper/2024/hash/1119587863e78451f080da2a768c4935-Abstract-Conference.html}
}

@article{huang2025liveavatar,
  title={Live Avatar: Streaming Real-time Audio-Driven Avatar Generation with Infinite Length},
  author={Huang, Yubo and Guo, Hailong and Wu, Fangtai and Wang, Weiqiang and Zhang, Shifeng and Huang, Shijie and Gan, Qijun and Liu, Lin and Zhao, Sirui and Chen, Enhong and Liu, Jiaming and Hoi, Steven},
  journal={arXiv preprint arXiv:2512.04677},
  year={2025},
  url={https://arxiv.org/abs/2512.04677}
}

@inproceedings{zou2026hiar,
  title={{HiAR}: Efficient Autoregressive Long Video Generation via Hierarchical Denoising},
  author={Zou, Kai and Zheng, Dian and Liu, Hongbo and Hang, Tiankai and Liu, Bin and Yu, Nenghai},
  booktitle={European Conference on Computer Vision},
  pages={225--242},
  year={2026},
  organization={Springer},
  url={https://link.springer.com/chapter/10.1007/978-3-032-37026-6_13}
}

@article{zhou2026editstream,
  title={{EditStream}: A Unified Autoregressive Framework for Interactive Video Generation and Editing},
  author={Zhou, Yuqian and Zhou, Zhenghong and Wu, Zongze and Smith, Cameron and Zhang, Richard and Luo, Jiebo and Shechtman, Eli and Lin, Zhe},
  journal={arXiv preprint arXiv:2608.21424},
  year={2026},
  url={https://arxiv.org/abs/2608.21424}
}

@manual{nvidia2026ampere,
  title={{NVIDIA Ampere GPU Architecture Tuning Guide}},
  author={{NVIDIA}},
  year={2026},
  note={CUDA 13.4, Secs.~1.4.1.2--1.4.1.3. Accessed September 21, 2026},
  url={https://docs.nvidia.com/cuda/ampere-tuning-guide/index.html}
}

@manual{nvidia2026cutlass,
  title={{CUTLASS}: Efficient {GEMM} in {CUDA}},
  author={{NVIDIA}},
  year={2026},
  note={Software Pipelining section. Accessed September 21, 2026},
  url={https://docs.nvidia.com/cutlass/latest/media/docs/cpp/efficient_gemm.html}
}

@inproceedings{carion2025sam3,
  title={{SAM} 3: Segment Anything with Concepts},
  author={Carion, Nicolas and Gustafson, Laura and Hu, Yuan-Ting and Debnath, Shoubhik and Hu, Ronghang and Suris Coll-Vinent, Didac and Ryali, Chaitanya and Alwala, Kalyan Vasudev and Khedr, Haitham and Huang, Andrew and Lei, Jie and Ma, Tengyu and Guo, Baishan and Kalla, Arpit and Marks, Markus and Greer, Joseph and Wang, Meng and Sun, Peize and R\"{a}dle, Roman and Afouras, Triantafyllos and Mavroudi, Effrosyni and Xu, Katherine and Wu, Tsung-Han and Zhou, Yu and Momeni, Liliane and Hazra, Rishi and Ding, Shuangrui and Vaze, Sagar and Porcher, Francois and Li, Feng and Li, Siyuan and Kamath, Aishwarya and Cheng, Ho Kei and Dollar, Piotr and Ravi, Nikhila and Saenko, Kate and Zhang, Pengchuan and Feichtenhofer, Christoph},
  booktitle={International Conference on Learning Representations},
  pages={138846--138923},
  year={2026},
  url={https://proceedings.iclr.cc/paper_files/paper/2026/hash/e0982cbc81401df3430ee1ff780dc7a2-Abstract-Conference.html}
}

@inproceedings{caron2021dino,
  title={Emerging Properties in Self-Supervised Vision Transformers},
  author={Caron, Mathilde and Touvron, Hugo and Misra, Ishan and J{\'e}gou, Herv{\'e} and Mairal, Julien and Bojanowski, Piotr and Joulin, Armand},
  booktitle={IEEE/CVF International Conference on Computer Vision},
  pages={9630--9640},
  year={2021},
  organization={IEEE}
}

@article{polyak2025moviegen,
  title={Movie Gen: A Cast of Media Foundation Models},
  author={Polyak, Adam and Zohar, Amit and Brown, Andrew and Tjandra, Andros and Sinha, Animesh and Lee, Ann and Vyas, Apoorv and Shi, Bowen and Ma, Chih-Yao and Chuang, Ching-Yao and Yan, David and Choudhary, Dhruv and Wang, Dingkang and Sethi, Geet and Pang, Guan and Ma, Haoyu and Misra, Ishan and Hou, Ji and Wang, Jialiang and Jagadeesh, Kiran and Li, Kunpeng and Zhang, Luxin and Singh, Mannat and Williamson, Mary and Le, Matt and Yu, Matthew and Singh, Mitesh Kumar and Zhang, Peizhao and Vajda, Peter and Duval, Quentin and Girdhar, Rohit and Sumbaly, Roshan and Rambhatla, Sai Saketh and Tsai, Sam and Azadi, Samaneh and Datta, Samyak and Chen, Sanyuan and Bell, Sean and Ramaswamy, Sharadh and Sheynin, Shelly and Bhattacharya, Siddharth and Motwani, Simran and Xu, Tao and Li, Tianhe and Hou, Tingbo and Hsu, Wei-Ning and Yin, Xi and Dai, Xiaoliang and Taigman, Yaniv and Luo, Yaqiao and Liu, Yen-Cheng and Wu, Yi-Chiao and Zhao, Yue and Kirstain, Yuval and He, Zecheng and He, Zijian and Pumarola, Albert and Thabet, Ali and Sanakoyeu, Artsiom and Mallya, Arun and Guo, Baishan and Araya, Boris and Kerr, Breena and Wood, Carleigh and Liu, Ce and Peng, Cen and Vengertsev, Dimitry and Schonfeld, Edgar and Blanchard, Elliot and Juefei-Xu, Felix and Nord, Fraylie and Liang, Jeff and Hoffman, John and Kohler, Jonas and Fire, Kaolin and Sivakumar, Karthik and Chen, Lawrence and Yu, Licheng and Gao, Luya and Georgopoulos, Markos and Moritz, Rashel and Sampson, Sara K. and Li, Shikai and Parmeggiani, Simone and Fine, Steve and Fowler, Tara and Petrovic, Vladan and Du, Yuming},
  journal={arXiv preprint arXiv:2410.13720},
  year={2024},
  url={https://arxiv.org/abs/2410.13720}
}

@inproceedings{gu2026anyflow,
  author    = {Yuchao Gu and Guian Fang and Yuxin Jiang and Weijia Mao and Song Han and Han Cai and Mike Zheng Shou},
  title     = {{AnyFlow}: Any-Step Video Diffusion Model with On-Policy Flow Map Distillation},
  booktitle = {Computer Vision - {ECCV} 2026 - 19th European Conference, Malm{\"o}, Sweden, September 8-12, 2026, Proceedings, Part {LII}},
  series    = {Lecture Notes in Computer Science},
  volume    = {17052},
  pages     = {163--181},
  publisher = {Springer},
  year      = {2026},
  doi       = {10.1007/978-3-032-37235-2_10},
  url       = {https://link.springer.com/chapter/10.1007/978-3-032-37235-2_10}
}

@misc{huang2026solarwm,
  title  = {{SolarWM}: Open Data and Scalable Training for Long-Horizon Video World Models},
  author = {Junchao Huang and Guian Fang and Shengju Qian and Xianghao Kong and Zhuoran Zhao and Wei Huang and Yihua Du and Zixin Zhang and Justin Cui and Yuchao Gu and Yukang Chen and Xinting Hu and Tianyu He and Shaoshuai Shi and Zhuotao Tian and Xin Wang and Mike Zheng Shou and Li Jiang},
  year   = {2026},
  eprint = {2609.02886},
  archivePrefix = {arXiv},
  primaryClass  = {cs.CV},
  url    = {https://arxiv.org/abs/2609.02886}
}

@misc{zhao2026maskforcing,
  title  = {{Mask Forcing}: Improving Autoregressive Video Diffusion Distillation via Dual-Noise Masking Rollout},
  author = {Zhuoran Zhao and Shengju Qian and Tongtong Liang and Xianghao Kong and Songchun Zhang and Junchao Huang and Guian Fang and Xin Wang and Pan Hui and Anyi Rao},
  year   = {2026},
  eprint = {2609.09123},
  archivePrefix = {arXiv},
  primaryClass  = {cs.CV},
  url    = {https://arxiv.org/abs/2609.09123}
}

@article{wang2026memforest,
  title   = {{MemForest}: Efficient Agent Memory Management via {EventTree} Partitioning and Progressive Merging},
  author  = {Wang, Junxi and Sun, Te and Zhu, Jiayi and Zhang, Chen and Li, Siyuan and Liu, Xuyang and Wen, Zichen and Tu, Xiaobing and Ren, Jinkui and Zhang, Xiantao and Yuan, Ziqi and Zhang, Linfeng},
  journal = {arXiv preprint arXiv:2609.08273},
  year    = {2026},
  url     = {https://arxiv.org/abs/2609.08273}
}

@inproceedings{wang2026streammeco,
  title     = {{StreamMeCo}: Long-Term Agent Memory Compression for Efficient Streaming Video Understanding},
  author    = {Wang, Junxi and Sun, Te and Zhu, Jiayi and Li, Junxian and Xu, Haowen and Wen, Zichen and Hu, Xuming and Li, Zhiyu and Zhang, Linfeng},
  booktitle = {Findings of the Association for Computational Linguistics: {ACL} 2026},
  year      = {2026},
  pages     = {13234--13251},
  publisher = {Association for Computational Linguistics},
  doi       = {10.18653/v1/2026.findings-acl.647},
  url       = {https://aclanthology.org/2026.findings-acl.647/}
}

@article{li2026forgewm,
  title   = {{ForgeWM}: Progressive Causal Training for Few-Step Action-Conditioned Video World Models},
  author  = {Li, Xinye and Lin, Lingshuai and Wang, Lei and Zhang, Liuzhou and Cui, Jialin and Li, Qingshan and Wang, Guanchu and Liu, Qingbin and Chen, Xi and Bian, Jiang and Lam, Wai},
  journal = {arXiv preprint arXiv:2608.14022},
  year    = {2026},
  url     = {https://arxiv.org/abs/2608.14022}
}

@article{su2026onewm,
  title   = {Can World and Action Modeling Benefit Each Other under Unified Causal Training?},
  author  = {Su, Yue and Lin, Bokai and Wang, Xinkai and Zhang, Chubin and Wu, Xiangnan and Zhang, Kuan and Jin, Weiyang and Yu, Jiwen and Li, Yiming and Lu, Cewu and Yang, Lixin and Liu, Xihui},
  journal = {arXiv preprint arXiv:2610.21095},
  year    = {2026},
  url     = {https://arxiv.org/abs/2610.21095}
}

@article{zhao2026minwm,
  title   = {{minWM}: A Full-Stack Open-Source Framework for Real-Time Interactive Video World Models},
  author  = {Zhao, Min and Zhu, Hongzhou and Yan, Bokai and Zhou, Zihan and Chen, Yimin and Wang, Honglie and Sun, Wenqiang and Fang, Zhengwei and Xun, Zizheng and Li, Zihao and Zheng, Kaiwen and He, Guande and Yang, Xiao and Li, Chongxuan and Bao, Fan and Zhu, Jun},
  journal = {arXiv preprint arXiv:2605.30263},
  year    = {2026},
  url     = {https://arxiv.org/abs/2605.30263}
}

@article{zhu2026sanawm,
  title   = {{SANA-WM}: Efficient Minute-Scale World Modeling with Hybrid Linear Diffusion Transformer},
  author  = {Zhu, Haoyi and Liu, Haozhe and Zhao, Yuyang and Ye, Tian and Chen, Junsong and Yu, Jincheng and He, Tong and Han, Song and Xie, Enze},
  journal = {arXiv preprint arXiv:2605.15178},
  year    = {2026},
  url     = {https://arxiv.org/abs/2605.15178}
}

@inproceedings{kodaira2025streamdiffusion,
  title     = {{StreamDiffusion}: A Pipeline-level Solution for Real-Time Interactive Generation},
  author    = {Akio Kodaira and Chenfeng Xu and Toshiki Hazama and Takanori Yoshimoto and Kohei Ohno and Shogo Mitsuhori and Soichi Sugano and Hanying Cho and Zhijian Liu and Masayoshi Tomizuka and Kurt Keutzer},
  booktitle = {Proceedings of the IEEE/CVF International Conference on Computer Vision},
  year      = {2025},
  pages     = {12371--12380},
  url       = {https://openaccess.thecvf.com/content/ICCV2025/papers/Kodaira_StreamDiffusion_A_Pipeline-level_Solution_for_Real-Time_Interactive_Generation_ICCV_2025_paper.pdf}
}

@inproceedings{feng2026streamdiffusionv2,
  title     = {{StreamDiffusionV2}: A Streaming System for Dynamic and Interactive Video Generation},
  author    = {Tianrui Feng and Zhi Li and Shuo Yang and Haocheng Xi and Muyang Li and Xiuyu Li and Lvmin Zhang and Keting Yang and Kelly Peng and Song Han and Maneesh Agrawala and Kurt Keutzer and Akio Kodaira and Chenfeng Xu},
  booktitle = {Proceedings of Machine Learning and Systems},
  volume    = {8},
  year      = {2026},
  url       = {https://proceedings.mlsys.org/paper_files/paper/2026/hash/698cfaf72a208aef2e78bcac55b74328-Abstract-Conference.html}
}

@article{xia2026slackserve,
  title   = {Adaptive Resource Management and Quality Control for Streaming Video Generation},
  author  = {Yifei Xia and Hao Yuan and Suhan Ling and Haoran Sun and Hanke Zhang and Xupeng Miao and Fangcheng Fu and Bin Cui},
  journal = {arXiv preprint arXiv:2606.15319},
  year    = {2026},
  url     = {https://arxiv.org/abs/2606.15319}
}

@misc{physis2026physisserve,
  title        = {{PhysisServe}: Efficient Autoregressive Denoising Serving with Fine-Grained Stage-wise Pipelining},
  author       = {{Physis AI Infra Team}},
  howpublished = {Technical blog},
  year         = {2026},
  month        = oct,
  url          = {https://physis-ai.com/blog/physisserve/en}
}

@article{chen2026codeworldmodel,
  title   = {{Code World Model}: Coding Agent as World Brain},
  author  = {Yiwen Chen and Guosheng Lin and Chi Zhang},
  journal = {arXiv preprint arXiv:2608.25927},
  year    = {2026},
  url     = {https://arxiv.org/abs/2608.25927}
}

@article{zhang2026matrixgame4,
  title   = {{Matrix-Game 4.0}: Harnessing Video Models for Real-Time Playable Game Worlds},
  author  = {Jihai Zhang and Runjia Qian and Kaiqin Kong and Biao Jiang and Puyi Wang and Siru Chen and Tianle Li and Guanjie Chen and Junyi Wu and Chubin Chen and Yuanhang Li and Kaichen Huang and Haobo Zhang and Yangguang Li and Wei Yu and Guibin Chen and Yu Cheng and Yahui Zhou},
  journal = {arXiv preprint arXiv:2610.27999},
  year    = {2026},
  url     = {https://arxiv.org/abs/2610.27999}
}

@misc{streamrenderh3_2026,
  title        = {{StreamRender-H3}: Interactive Coding Worlds with Real-time Streaming Diffusion Render},
  author       = {Yanzuo Lu and Tian Ye and Yitong Li and Shuchen Xue and Haozhe Liu and Song Han and Enze Xie},
  howpublished = {Research blog},
  year         = {2026},
  url          = {https://nvlabs.github.io/Sana/Sol-Engine/StreamRender-H3/}
}

@misc{fang2026solarrenderer,
  title  = {{SolarRenderer}: Few-Step Generative Reference Rendering},
  author = {Guian Fang and Shengju Qian and Dongyue Lu and Junchao Huang and Lutao Jiang and Xin Wang and Yiren Song and Li Jiang and Mike Zheng Shou},
  year   = {2026},
  eprint = {2610.28630},
  archivePrefix = {arXiv},
  primaryClass  = {cs.CV},
  url    = {https://arxiv.org/abs/2610.28630}
}

@article{zhang2026vidus1,
  title   = {{Vidu S1}: A Real-Time Interactive Video Generation Model},
  author  = {Jintao Zhang and Kai Jiang and Jintao Chen and Xu Wang and Yang Luo and Yuji Wang and Dechuang Chen and Jungang Li and Chengyang Ye and Marco Chen and Hongzhou Zhu and Min Zhao and Yuxuan Jiang and Zhengkun Huang and Chendong Xiang and Kaiwen Zheng and Haoxu Wang and Xiaohang Wang and Qi Jia and Xin Chen and Yimin Chen and Youhe Jiang and Fangcheng Fu and Zhijie Deng and Fan Bao and Jianfei Chen and Jun Zhu},
  journal = {arXiv preprint arXiv:2607.03118},
  year    = {2026},
  url     = {https://arxiv.org/abs/2607.03118}
}

@article{zhang2026vidus2,
  title   = {{Vidu S2}: Real-Time Interactive, Editable, and Spatial Video Generation},
  author  = {Jintao Zhang and Kai Jiang and Jintao Chen and Xu Wang and Deyuan Liu and Jungang Li and Dechuang Chen and Ming Lin and Jingjiang Zhou and Haopeng Jin and Qi Jia and Xiaohang Wang and Yaole Wang and Zhanqiang Zhang and Ran Li and Zhengkun Huang and Shuyue Xiong and Yuji Wang and Zikun Dai and Hui He and Yang Luo and Mang Ning and Weiqi Feng and Chengyang Ye and Xinyue Lin and Min Zhao and Hongzhou Zhu and Hengkai Tan and Zeyuan Wang and Chendong Xiang and Kaiwen Zheng and Zhijie Deng and Fan Bao and Jianfei Chen and Jun Zhu},
  journal = {arXiv preprint arXiv:2609.11638},
  year    = {2026},
  url     = {https://arxiv.org/abs/2609.11638}
}

@article{streameval2026,
  title   = {{StreamEval}: Evaluating Interactive Streaming Video Generation Models},
  author  = {{StreamEval Team}},
  journal = {arXiv preprint arXiv:2610.28402},
  year    = {2026},
  url     = {https://arxiv.org/abs/2610.28402}
}

\clearpage
\appendix

\section{Training Implementation Details}
\label{sec:training-appendix}

\paragraph{Noise schedules and score networks.}
The 4-step schedule is $[1000,750,500,250]$ and the 5-step schedule is $[1000,800,600,400,200]$, before timestep warping with shift factor 5. Both use three latent frames per chunk and preserve their respective schedules through all stages and inference. The frozen DMD teacher is bidirectional Wan2.1-T2V-14B. The trainable fake-score model starts from bidirectional Wan2.1-T2V-1.3B and does not share weights with the causal student. It learns to denoise noisy, detached student predictions. Stage transitions transfer raw student weights and restart the fake-score model, optimizer, and EMA states. Evaluation uses the resulting EMA student.

\paragraph{Off-policy targets and perceptual loss.}
Stage 2 uses 2048 training and 128 validation teacher examples generated by Wan2.1-T2V-14B with 50 denoising steps, guidance scale 3, and timestep shift 5. Each example contains 21 latent frames. The off-policy implementation samples $S$ consecutive chunks ($3S$ latent frames), constructs student K/V from preceding clean teacher chunks, and re-noises the sampled chunks to form a set of chunks at staggered noise levels. DMD and LPIPS are applied to the same clean prediction, with LPIPS weight 0.25 for both model versions. The fake-score update uses student predictions from these teacher-derived states.

For LPIPS, latent spatial resolution is bilinearly reduced from $60\times104$ to $30\times52$, then decoded by a frozen Wan VAE to $240\times416$ RGB frames. Every third decoded frame is evaluated. The 4-step training input contains 12 latent frames and decodes to 45 frames, giving 15 perceptual comparisons. The 5-step training input contains 15 latent frames and decodes to 57 frames, giving 19 comparisons. Decoded values are clamped to $[-1,1]$, and VGG-based LPIPS is averaged over the sampled frames.

\paragraph{Optimization and updates.}
The student and fake-score learning rates are $1.5\times10^{-6}$ and $4\times10^{-7}$, with Adam moments $(0,0.999)$. EMA uses decay 0.99 and starts after 200 training iterations. The training used eight workers with one sample each and no gradient accumulation, giving an effective batch size of eight. For the three-stage 4-step model, the 3000/1000/2000 training-loop iterations correspond to 600/1000/400 student updates and 3000/1000/2000 fake-score updates. The 5-step model starts from released RF weights and proceeds directly to Stage 2 (Sec.~\ref{sec:training}), so the Stage 1 iteration and update counts apply only to the 4-step model. Stage 1 and Stage 3 update the student once per five fake-score updates, whereas Stage 2 updates both \mbox{models each iteration}.

\paragraph{Rollout visibility.}
The Stage 3 sampler selects RF, WF, or SF for the same student with equal probability. The WF path reuses the RF rollout implementation with chunk-wise causal attention to advance the wavefront: clean history and the same or earlier active chunks are visible, while later active chunks are masked. This matches Eq.~\eqref{eq:wf-mask}. The RF and SF paths remain complementary training distributions.

\FloatBarrier

\section{Runtime Implementation and Analysis}
\label{sec:runtime-appendix}

\subsection{Process ownership and reference execution}

WaveRT uses $S$ denoising processes and one store process, each owning a GPU and a full DiT replica. A dedicated group of $P_{\mathrm{VAE}}$ GPUs owns contiguous decoder partitions, giving $S+1+P_{\mathrm{VAE}}$ GPUs. GPU allocations list denoising + Store + VAE: $4+1+3$ for 4-step and $5+1+2$ for 5-step WaveRT, both totaling eight GPUs. The store process performs a full clean-state cache refresh, rather than an \mbox{additional denoising update}.

The local forward uses SGLang Diffusion for loading, conditioning, scheduling, dense operators, and cross-attention. WaveRT interposes after self-attention QKV projection and RoPE to publish K/V and construct the legal context. It owns the temporal pages and replaces the base runtime's unused large self-attention cache with a placeholder. The execution tests use BF16/FA3 as the reference backend. The approximate compute modes replace self-attention with SageAttention and optionally use W8A8 linear layers, while retaining FA3 for cross-attention.

The synchronized reference gathers every rank's layer K/V, including store:
\begin{equation}
 \bigl((K_{k,r}^{\ell},V_{k,r}^{\ell})\bigr)_{r=0}^{S}
 =\operatorname{AllGather}_{r=0}^{S}(K_{k,r}^{\ell},V_{k,r}^{\ell}).
 \label{eq:sync-pool}
\end{equation}
Idle ranks provide dummy payloads during fill/drain, but their slots never enter attention. Each active rank keeps all query heads. The RF path reads every active denoising slot, whereas WF selects the context in Eq.~\eqref{eq:wf-context}. Synchronized and asynchronous WF share the same context rules. Native RF comparison additionally changes the visibility graph and requires matching history and positions.

\subsection{K/V publication and buffer lifetime}
\label{sec:systems-analogy}

At initialization, consumers allocate remote K/V slots and device-side generation flags indexed by producer, round buffer, and layer, then exchange IPC memory handles. After rank $j$ produces layer $\ell$ K/V, its communication stream waits on the producer event, copies the payload to each active consumer $s<j$, and publishes the generation flag. A consumer's compute stream waits for the matching flag immediately before attention. All legal cleaner ranks, including nonadjacent chunks and active store, \mbox{must be ready}.

This produce--publish--wait-before-use order follows buffered software-pipeline principles~\citep{nvidia2026ampere,nvidia2026cutlass}, applied here to newly computed inter-GPU K/V rather than prefetched static tiles. Causal visibility permits producer lead but does not guarantee it. WaveRT enforces the order explicitly because ordinary CTA scheduling supplies no such guarantee. The CUDA IPC implementation uses \texttt{cuMemcpyDtoDAsync}, whose peer copies may execute through an SM kernel. 

The paged path writes remote segments directly into their final attention locations:
\begin{equation}
 \mathrm{KPage},\mathrm{VPage}\in
 \mathbb R^{2\times L\times N_{\max}\times n_{\mathrm{head}}\times d_{\mathrm{head}}}.
 \label{eq:paged-layout}
\end{equation}
The leading dimension double-buffers rounds, and $N_{\max}$ bounds context tokens. Consumers fill anchor, retained history, and self slots, while producers fill legal remote slots. Attention reads the resulting ordered prefix without per-layer remote \texttt{torch.cat}. Here, ``paged'' means fixed-slot placement and does not eliminate local/history copies. Outgoing copies complete before the round boundary permits reuse, and generation flags distinguish successive uses of a slot.

\subsection{Round schedule and cache semantics}

Algorithm~\ref{alg:wave-runtime} implements the coordinates and active set defined in Sec.~\ref{sec:wf}. It runs $N_{\mathrm{chunk}}+S$ rounds, including fill/drain. The latent All-Gather at each round boundary remains in the asynchronous implementation and is distinct from the removed per-layer K/V collective.

\begin{algorithm}[t]
\caption{WaveRT round schedule (one request, $N_{\mathrm{chunk}}$ chunks)}
\label{alg:wave-runtime}
\begin{algorithmic}[1]
\Require $S$ denoising ranks, store rank $S$, identical weights and fixed scheduler
\State Initialize empty anchor/history and generation-indexed K/V buffers
\For{$k=0,\ldots,N_{\mathrm{chunk}}+S-1$}
  \State On each rank $r$, execute the following concurrently with $i=k-r$
  \If{$r<S$ and $0\leq i<N_{\mathrm{chunk}}$}
    \State Rank 0 admits $Z_i^{(0)}$, while other ranks use the previous tick's latent
    \For{$\ell=1,\ldots,L$}
      \State Produce local Q/K/V and publish K/V to active ranks $s<r$
      \State Wait for legal cleaner-rank and store pages for $(k,\ell)$
      \State Apply attention with Eq.~\eqref{eq:wf-context}, then finish the layer
    \EndFor
    \State Apply the stage transition with noise keyed by chunk and stage
  \ElsIf{$r=S$ and $0\leq i<N_{\mathrm{chunk}}$}
    \State Refresh K/V from clean $B_i$ and publish each layer's K/V to active denoising ranks
    \State Enqueue $B_i$ for the streaming VAE
  \EndIf
  \State Drain outgoing copies and promote the current store page into history
  \State All ranks participate in latent All-Gather, after which each receiver takes its predecessor's output
\EndFor
\end{algorithmic}
\end{algorithm}

Context order is exactly Eq.~\eqref{eq:wf-context}: anchor, bounded clean history, live store, self, then cleaner denoising slots in increasing stage order. The live store page becomes retained history after the round, avoiding a second steady-state broadcast. The first clean chunk serves as a persistent anchor, with its unrotated key retained so RoPE can be reapplied when the bounded window shifts. Other pages reuse their stored RoPE-applied keys.

Store processes $B_{k-S}$ at the configured \texttt{context\_noise} timestep, discards its prediction, and keeps each layer's K/V. This timestep is specified independently of $S$. Store attends to its anchor, retained history, and self. Its history budget reserves one query chunk, whereas denoising ranks reserve $S$ chunks, so their prefix lengths can differ despite sharing the same clean-page list. Refresh timestep, prefix budgets, positions, and promotion order are part of the fixed semantics.

Conditioning, attention pages, and VAE convolution caches are request-local. The resident launcher handles one queued request at a time.

\subsection{Output cadence, work, and fill/drain}
\label{sec:execution-cost}

Let $T_s$ be stage $s$'s local service time and $T_{\mathrm{store}}$ the clean-refresh time, measured for a specified graph, context, and backend. Sequential generation has period
\begin{equation}
 T_{\mathrm{serial}}^{(S)}=\sum_{s=0}^{S-1}T_s+T_{\mathrm{store}},
 \qquad T_{\mathrm{serial}}^{(1)}=T_0+T_{\mathrm{store}}.
 \label{eq:serial-cadence}
\end{equation}
With $S+1$ stage-owned replicas, the filled wave period satisfies
\begin{equation}
 T_{\mathrm{wave}}^{(S)}\geq\max\{T_0,\ldots,T_{S-1},T_{\mathrm{store}}\},
 \label{eq:wave-cadence}
\end{equation}
with equality only if communication and synchronization are fully covered. This permits an $S$-update pipeline to emit faster than serial one-step inference, but uses more replicas and does not reduce work or first-output latency. Across $N_{\mathrm{chunk}}$ chunks it still executes $N_{\mathrm{chunk}}(S+1)$ full forwards. Structural slot occupancy is
\begin{equation}
 \eta_{\mathrm{slot}}=\frac{N_{\mathrm{chunk}}}{N_{\mathrm{chunk}}+S}.
 \label{eq:slot-occupancy}
\end{equation}
All replicas are active only for $S\leq k<N_{\mathrm{chunk}}$, so requests with $N_{\mathrm{chunk}}\leq S$ never fill the pipeline. The 399-latent-frame workload has 133 chunks, giving $133/137=97.1\%$ slot occupancy at $S=4$. First-output latency includes all denoising and store stages, VAE pipeline \mbox{fill, and delivery}.

\subsection{Streaming VAE partitioning}
\label{sec:streaming-vae}

Completed latents enter the decoder queue. Each latent-frame microbatch traverses contiguous decoder stages in order, preserving per-request causal-convolution caches. Profiling $M_{\mathrm{VAE}}$ ordered units at representative streaming shapes gives service times $\tau_m^{\mathrm{VAE}}$. Dynamic programming chooses $P_{\mathrm{VAE}}$ partitions by
\begin{equation}
 \min_{0=b_0<\cdots<b_{P_{\mathrm{VAE}}}=M_{\mathrm{VAE}}}
 \max_{1\leq p\leq P_{\mathrm{VAE}}}
 \sum_{m=b_{p-1}+1}^{b_p}\tau_m^{\mathrm{VAE}}.
 \label{eq:partition-objective}
\end{equation}
This balances estimated local service, including costs missed by unit counts or FLOPs. Inter-stage delivery must also fit the arrival interval. In per-latent-chunk units, the steady decoded-output interval is~approximately
\begin{equation}
 T_{\mathrm{emit}}^{\mathrm{E2E}}\approx
 \max\!\left(T_{\mathrm{round}}^{\mathrm{WP}},
       \max_p T_{\mathrm{VAE},p}^{\mathrm{chunk}}\right)
       +T_{\mathrm{delivery}}^{\mathrm{exposed}},
 \label{eq:e2e-max}
\end{equation}
where decoder service amortizes all frame microbatches in one chunk, and the last term counts residual delivery not already covered by overlap or the service terms. Slower decoding grows queues or triggers bounded-buffer backpressure. The model excludes fill/drain and file encoding. Tab.~\ref{tab:mechanism-diagnostics} reports profiling, while Fig.~\ref{fig:vae-pipeline} illustrates the schedules.
\subsection{Same-graph semantic preservation}
\label{sec:semantic-correctness}

Wave Forcing changes the model graph relative to RF. Once the WF graph is fixed, synchronized and asynchronous execution can be compared at the operator level.

\begin{proposition}[Same-graph scheduling]
Fix the model parameters, denoising schedule, initial noise, per-update re-noising draws, text conditioning, positional embeddings, and history/store rules. Assume each local operator evaluates the same mathematical function and every attention operation receives the ordered K/V context in Eq.~\eqref{eq:wf-context}, after the corresponding K/V production and delivery have completed. Then dependency-respecting asynchronous execution and synchronized execution produce the same mathematical latent sequence.
\end{proposition}

\begin{proof}\

At the first tick, both executions start from identical inputs and empty history. At each layer, local Q/K/V projections depend only on that rank's current hidden states and fixed conditioning. The synchronized execution gathers all projections before consuming them. The asynchronous execution may publish and consume them at different times, but the readiness condition guarantees the same legal K/V values and sequence order at attention. Consequently each attention result and the remaining local layer outputs agree. Induction over layers gives identical clean-latent predictions, from which the fixed scheduler and noise draws produce identical stage outputs. Store receives the same clean latent, runs the same cache-refresh computation, and promotes the same pages after the tick. Induction over ticks establishes identical histories and emitted latents through fill, steady state, and drain.
\end{proof}


\subsection{Timing contract and multiple producers}
\label{sec:timing-contract}

On a common time base, let $t_v^{\mathrm{local}}$ be consumer $v=C_{k,s,\ell}$'s local-input/compute-stream ready time before remote waits. For each required edge $e$, set $t_e^{\mathrm{need}}=t_v^{\mathrm{local}}$ and $t_e^{\mathrm{arrive}}=t_e^{\mathrm{ready}}+\tau_e$. Eq.~\eqref{eq:slack} then gives
\begin{equation}
 t_v^{\mathrm{start}}
 =\max\!\left(t_v^{\mathrm{local}},
              \max_{e\in\operatorname{In}_{\mathrm{KV}}(v)}t_e^{\mathrm{arrive}}\right),
 \qquad
 w_v=t_v^{\mathrm{start}}-t_v^{\mathrm{local}}
     =\max_{e\in\operatorname{In}_{\mathrm{KV}}(v)}w_e.
 \label{eq:consumer-wait}
\end{equation}
The incoming set includes all legal cleaner ranks and active store. When it is empty, $t_v^{\mathrm{start}}=t_v^{\mathrm{local}}$ and $w_v=0$. Earlier waits already affect these timestamps. Round critical paths additionally include transfer serialization, contention, and the boundary collective. Summing concurrent durations would double-count overlap. Measuring this contract requires correlated timestamps across calibrated device clocks, not just aggregate rank times.

\subsection{Bounded attention storage and communication}
\label{sec:storage-communication}

The two K/V arrays in Eq.~\eqref{eq:paged-layout} have fixed capacity for anchor, bounded history, live store, and at most $S$ denoising chunks, so attention-page storage is independent of generated video length. This bound excludes model weights and other request state. Explicitly retained video context is limited to the anchor and recent history.

For direct publication at steady state, denoising rank $j$ sends its K/V chunk to $j$ noisier ranks, for $S(S-1)/2$ payload destinations among denoising stages per layer. Store adds $S$ destinations within the full wavefront. This quadratic count in $S$ covers logical payload copies, while physical link traffic depends on transport and topology. The practical gain is reduced exposed communication under the timing condition in Eq.~\eqref{eq:slack}.

\paragraph{Index conventions.}
Chunk, denoising-stage, and rank indices are zero-based, whereas Transformer layers and VAE partitions are one-based. Diagram cells are tasks, whereas $Z_i^{(s)}$ is an input state. A round is a logical unit with variable wall time. Tab.~\ref{tab:notation} summarizes the graph-specific symbols, with local timing and shape variables defined alongside their equations.
\begin{table}[htbp]
\centering\small
\caption{Graph and runtime notation. The complete wavefront includes store, whereas the denoising \mbox{subset does not}.}
\label{tab:notation}
\begin{tabularx}{\linewidth}{@{}lX@{}}
\toprule
\tabhead{Symbol} & \tabhead{Meaning} \\
\midrule
$N_{\mathrm{chunk}},i$ & $N_{\mathrm{chunk}}$ is the number of requested chunks, and $i$ indexes chunks with $0\leq i<N_{\mathrm{chunk}}$. \\
$S,s,\nu_s$ & $S$ is the number of denoising updates, $s$ indexes denoising stages with $0\leq s<S$, and $\nu_s$ is the diffusion timestep. Store occupies execution stage $s=S$. \\
$k,r$ & $k$ indexes forward rounds, and $r$ indexes diffusion ranks with $0\leq r\leq S$, where rank $S$ is Store. \\
$Z_i^{(s)},B_i$ & $Z_i^{(s)}$ is the state after $s$ updates, and $B_i=Z_i^{(S)}$ denotes the clean endpoint. \\
$\mathcal A_k,\mathcal F_k$ & active denoising-stage indices and their input states \\
$\mathcal W_k$ & complete wavefront state, including the clean input to store when active \\
$\mathcal C_{k,s}^{\mathrm{WF},\ell}$ & Ordered K/V context, with $k=i+s$ relating the round-based and chunk-based notation. \\
$A_k^\ell,H_k^\ell,U_k^\ell$ & anchor, bounded clean-history, and live store K/V sequences \\
$P_{k,s,\ell},C_{k,s,\ell}$ & $P_{k,s,\ell}$ denotes Q/K/V production, and $C_{k,s,\ell}$ denotes attention and the remaining block tasks. \\
\bottomrule
\end{tabularx}
\end{table}

\FloatBarrier

\section{Evaluation details}
\label{sec:evaluation-details}

\paragraph{VBench protocol.}
We follow the standard VBench benchmark~\citep{huang2024vbench} with the original prompts, normalization, and weights, and scale scores to 0--100. For detailed metric definitions, we refer readers to the VBench paper. All metrics are computed with the official standardized evaluation scripts.

\paragraph{VBench-Long protocol.}
We report the VBench-Long Quality score and Dynamic Degree, scaled to 0--100. Each method uses the same 200 MovieGen prompts and seed 42, generating 399 latent frames and saving 1593 frames (about 100 seconds) at 16 FPS per video. Causal Forcing (Full Attn.) uses the EMA generator of the released \texttt{chunkwise/longvideo.pt} checkpoint with its native 4-step schedule and full attention, for both standard VBench and VBench-Long.

\paragraph{Quality-generation setup.}
Rolling Forcing and Causal Forcing are evaluated in their own code repositories. RF with the causal mask and Wave Forcing use WaveRT with causal KV context, paged exchange, and a time-partitioned VAE. The 4-step and 5-step models use 4+1+3 and 5+1+2 GPUs, respectively, ordered as denoising + Store + VAE. Checkpoint quality is evaluated with SDPA for standard inference and SageAttention for fast inference. Execution performance is measured separately using FA3 and dummy DiT weights.

\subsection{Long-video stability diagnostics}
\label{sec:diagnostic-implementation}

\paragraph{Prompt subset.}
Both diagnostics require a principal subject named in the prompt. A fixed text rule first removes, for every method, group scenes, scenery, first-person views without a stable visible subject, ambiguous principal entities, and prompts that ask the subject to disappear or transform. Of the 200 MovieGen prompts, 131 pass this rule. We then sample 100 of these prompts once at random and hold that subset fixed across methods, ablation studies, and both diagnostics. This subject-focused subset is distinct from the full 200-prompt VBench-Long evaluation.

\paragraph{Subject Preservation.}
We measure the prompt's principal-subject visibility.

SAM3 tracks the text-specified entity at 8 FPS, giving 797 sampled frames per video. The initial binding is the largest detection in the first eight frames, and masked DINO~\citep{caron2021dino} features decide which later detections share that identity. Let $A_t$ be the current mask area and $A_0$ the median mask area of the bound identity over the eight frames that begin at this binding. An associated box within $1\%$ of the image border has visibility weight $\min(1,A_t/(0.18A_0))$, while any other associated detection has weight 1. A new instance elsewhere keeps its own track while the recognized subject still has weight at least $0.5$ and mask area at least $0.18A_0$. Frames without an associated detection have weight 0. The video score is 100 times the mean weight over all sampled frames, including frames before the binding and frames without support. Videos that never receive an initial binding, and videos whose binding was reviewed as incorrect, are left unscored.

\paragraph{Camera Drift.}
Camera Drift measures persistent background motion with the subject moving toward an image boundary. We use the union of SAM3 detections for the text-specified entity at 8 FPS and point trajectories seeded outside those masks at 4 FPS. All distances are normalized by the image diagonal. For each image edge $e$, let $g_e$ and $q_e$ be the 5th and 90th percentiles of mask-pixel distances to that edge. Foreground outward velocity is the decrease in $q_e$ divided by the 0.25-second pair duration. Background velocity is the component-wise median of per-cell median track velocities over supported spatial cells, projected onto the same outward axis. We apply $z(v)=\operatorname{sign}(v)\max(|v|-0.002,0)$ to both velocities.

Within the preceding five seconds, we use pairs with observed foreground and background motion. For either corrected velocity sequence, define outward displacement $A=\max(0,0.25\sum_t v_t)$, path length $P=0.25\sum_t|v_t|$, and response $H=(A/P)A^2/(A^2+0.1^2)$, with $H=0$ when $P=0$. These give foreground and background factors $F_e$ and $B_e$. Edge proximity is $E_e=\max(0,1-g_e/0.1)$. The instantaneous risk is $\max_e E_eF_eB_e$ over edges with positive current foreground and background velocities, or zero if no edge qualifies. Thus, background motion with a stationary subject does not incur a drift penalty.

To retain evidence after an exit, we require one second of consecutive outward foreground and background motion, at least two native pixels of foreground displacement, a final $g_e$ within two native pixels of the boundary, and $q_e\leq0.1$. After one second of consecutive explicitly empty masks, we hold the exit's $E_eF_e$ fixed and update $B_e$ from current background observations, without backfilling the confirmation interval. Missing background tracks can be replenished by points consistent with the existing motion layer, but new points contribute only after tracking through a subsequent pair. This is a geometric association, not an independent semantic background label. A returning detection, uncertain mask, or scene cut ends the extension. Stopped or reversed background motion immediately gives zero risk and ends the extension after one consecutive observed second. Background gaps remain unscored, and later continuation is distinguished from uninterrupted support.

The video score is 100 times the mean risk over valid 0.25-second intervals, so lower is better. Startup intervals before five seconds, unavailable current measurements, and windows crossing a detected or unresolved cut remain unscored. Missing motion within a window is neither interpolated nor extrapolated. We average video scores equally over the prompt subset, requiring at least 20 scored seconds per video and using each method's own valid intervals.
\subsection{Complete performance measurements}
\label{sec:performance-tables}

Tables~\ref{tab:runtime-core} and~\ref{tab:transport-ablation} provide the WaveRT measurements plotted in Fig.~\ref{fig:systems-performance}a,b, including matched DiT/E2E rates in Tab.~\ref{tab:runtime-core}. Table~\ref{tab:cp-dit} reports DiT throughput for each model scale, backend, and parallel \mbox{strategy in Fig.~\ref{fig:systems-performance}c--f}.

\paragraph{Experiment groups.}
Tables~\ref{tab:runtime-core}, \ref{tab:transport-ablation}, and~\ref{tab:repeat-ranges} draw from the same fixed-three-repeat system sweep, reporting selected runtime configurations, pooled steady tick p50, and per-run ranges, respectively. Shared configurations report the same experimental runs. Table~\ref{tab:cp-dit} is a separate DiT-only parallel-strategy sweep. Comparisons must respect each table's step counts, GPU allocations, timing scopes, and repetition cohorts.

\begin{table}[H]
\centering\small
\setlength{\tabcolsep}{3pt}
\caption{Runtime throughput. WaveRT uses the WF graph and paged K/V unless marked Sync, while Baseline is single-GPU rolling inference. Both use BF16 with FA3 unless a kernel suffix is specified. GPU counts include denoising, Store, and VAE: 4+1+3 for 4-step and 5+1+2 for 5-step. Speedup is E2E relative to the same-scale, same-step baseline. All rows use dummy DiT weights and a real-weight VAE. Best values are determined within each model/step group using unrounded measurements. \mbox{DiT/E2E are paired}.}
\label{tab:runtime-core}
\label{tab:main-throughput}
\begin{tabular}{@{}l*{3}{c}ccc@{}}
\toprule
\tabhead{Configuration} & \tabhead{GPUs} & \tabhead{Tick p50} & \tabhead{Steady} & \tabhead{DiT} & \tabhead{E2E} & \tabhead{Speedup} \\
 & & \tabhead{(ms)} & \tabhead{(FPS)} & \tabhead{(FPS)} & \tabhead{(FPS)} & \\
\midrule
\multicolumn{7}{c}{\tabhead{\textit{1.3B, 4-step}}} \\
Baseline & 1 & -- & -- & 33.86 & 19.06 & $1.00\times$ \\
WaveRT (Sync) & 8 & 91.96 & 130.50 & 129.31 & 127.06 & $6.67\times$ \\
WaveRT & 8 & 78.88 & 152.13 & 147.58 & \tabbest{134.81} & $\tabbestmath{7.07\times}$ \\
WaveRT + Sage & 8 & 78.36 & 153.15 & \tabbest{147.78} & 134.79 & $7.07\times$ \\
WaveRT + Sage + W8A8 & 8 & \tabbest{76.05} & \tabbest{157.79} & 145.47 & 132.89 & $6.97\times$ \\
\midrule
\multicolumn{7}{c}{\tabhead{\textit{1.3B, 5-step}}} \\
Baseline & 1 & -- & -- & 29.12 & 17.46 & $1.00\times$ \\
WaveRT & 8 & 120.53 & 99.56 & \tabbest{104.82} & 96.46 & $5.52\times$ \\
WaveRT + Sage & 8 & \tabbest{120.52} & \tabbest{99.57} & 104.80 & \tabbest{96.48} & $\tabbestmath{5.53\times}$ \\
WaveRT + Sage + W8A8 & 8 & 120.59 & 99.51 & 103.49 & 95.15 & $5.45\times$ \\
\midrule
\multicolumn{7}{c}{\tabhead{\textit{14B, 4-step}}} \\
Baseline & 1 & -- & -- & 6.31 & 5.51 & $1.00\times$ \\
WaveRT (Sync) & 8 & 464.27 & 25.85 & 25.56 & 25.47 & $4.62\times$ \\
WaveRT & 8 & 413.80 & 29.00 & 28.74 & 28.64 & $5.20\times$ \\
WaveRT + Sage & 8 & 414.03 & 28.98 & 28.72 & 28.62 & $5.19\times$ \\
WaveRT + Sage + W8A8 & 8 & \tabbest{354.45} & \tabbest{33.86} & \tabbest{33.64} & \tabbest{33.50} & $\tabbestmath{6.08\times}$ \\
\bottomrule
\end{tabular}
\end{table}

\begin{table}[H]
\centering
\small
\setlength{\tabcolsep}{4.2pt}
\caption{WaveRT same-graph system ablations report pooled steady tick p50 (ms, lower is better) for the 4-step, 8-GPU (4+1+3) setup across three fixed repeats. Each row compares execution strategies under a fixed compute backend. All runs use dummy DiT weights and a real-weight VAE and measure \mbox{execution cost only}.}
\label{tab:transport-ablation}
\begin{tabular}{ll*{5}{c}}
\toprule
\tabhead{Backbone} & \tabhead{Compute} & \tabhead{Sync} & \tabhead{Overlap} & \tabhead{One-sided} & \tabhead{Staggered} & \tabhead{Paged} \\
\midrule
1.3B & BF16 & 91.956 & 85.730 & 81.669 & 81.953 & \tabbest{78.882} \\
1.3B & Sage & 91.441 & 84.103 & 80.934 & 81.062 & \tabbest{78.356} \\
1.3B & Sage+W8A8 & 88.714 & 81.595 & 78.030 & 78.081 & \tabbest{76.052} \\
\midrule
14B & BF16 & 464.270 & 449.689 & 425.907 & 425.188 & \tabbest{413.795} \\
14B & Sage & 467.526 & 439.988 & 426.215 & 425.601 & \tabbest{414.029} \\
14B & Sage+W8A8 & 407.215 & 379.576 & 365.922 & 366.697 & \tabbest{354.447} \\
\bottomrule
\end{tabular}
\end{table}

\begin{table}[htbp]
\centering\small
\setlength{\tabcolsep}{3.5pt}
\caption{DiT throughput across parallel strategies and compute backends, using dummy DiT weights with VAE disabled. Step counts and GPU counts are matched across methods within each row. FPS is the arithmetic mean of per-run rates, and best values are determined using unrounded measurements. WP/CP divides WP throughput by that of the faster CP baseline. Amber $\textcolor{tableamber}{\dagger}$ marks supplemental measurements whose repetition and stopping protocols are described in Appendix~\ref{sec:parallelism-analysis}.}
\label{tab:cp-dit}
\begin{tabular}{@{}llrrrrrr@{}}
\toprule
& & & & \multicolumn{3}{c}{\tabhead{DiT throughput (FPS) $\uparrow$}} & \\
\cmidrule(lr){5-7}
\tabhead{Scale} & \tabhead{Compute} & \tabhead{Steps} & \tabhead{GPUs} & \tabhead{Ulysses CP} & \tabhead{Ring CP} & \tabhead{WP} & \tabhead{WP/CP} \\
\midrule
1.3B & BF16 & 5 & 6 & 80.60 & 41.19 & \tabbest{150.09}\tabdiagmark & \tabgain{$1.862\times$}\tabdiagmark \\
1.3B & Sage & 5 & 6 & 84.01 & 56.99 & \tabbest{151.15} & \tabgain{$1.799\times$} \\
1.3B & Sage+W8A8 & 5 & 6 & 61.17 & 45.96 & \tabbest{154.75} & \tabgain{$2.530\times$} \\
\midrule
3B & BF16 & 5 & 6 & 75.38 & 38.20 & \tabbest{88.86} & \tabgain{$1.179\times$} \\
3B & Sage & 5 & 6 & 75.93 & 37.00 & \tabbest{89.26} & \tabgain{$1.176\times$} \\
3B & Sage+W8A8 & 5 & 6 & 56.15 & 36.04 & \tabbest{95.31} & \tabgain{$1.697\times$} \\
\midrule
5B & BF16 & 5 & 6 & 58.40 & 41.42 & \tabbest{68.33} & \tabgain{$1.170\times$} \\
5B & Sage & 5 & 6 & 63.13 & 30.37 & \tabbest{67.00} & \tabgain{$1.061\times$} \\
5B & Sage+W8A8 & 5 & 6 & 60.71 & 31.17 & \tabbest{75.44} & \tabgain{$1.243\times$} \\
\midrule
7B & BF16 & 5 & 6 & 49.89 & 38.96 & \tabbest{53.28} & \tabgain{$1.068\times$} \\
7B & Sage & 5 & 6 & 52.70\tabdiagmark & 25.75 & \tabbest{53.50}\tabdiagmark & \tabgain{$1.015\times$}\tabdiagmark \\
7B & Sage+W8A8 & 5 & 6 & 56.12 & 26.56 & \tabbest{61.55} & \tabgain{$1.097\times$} \\
\midrule
9B & BF16 & 5 & 6 & \tabbest{46.41}\tabdiagmark & 38.94 & 45.70\tabdiagmark & \tabloss{$0.985\times$}\tabdiagmark \\
9B & Sage & 5 & 6 & 45.26 & 22.73 & \tabbest{45.59} & \tabgain{$1.007\times$} \\
9B & Sage+W8A8 & 5 & 6 & 52.51 & 24.16 & \tabbest{53.84} & \tabgain{$1.025\times$} \\
\midrule
14B & BF16 & 4 & 5 & 26.72 & 23.90 & \tabbest{28.71} & \tabgain{$1.075\times$} \\
14B & Sage & 4 & 5 & 27.90 & 16.32 & \tabbest{28.72} & \tabgain{$1.029\times$} \\
14B & Sage+W8A8 & 4 & 5 & 30.98 & 17.16 & \tabbest{33.60} & \tabgain{$1.085\times$} \\
\bottomrule
\end{tabular}
\space
\end{table}

\begingroup
\setlength{\emergencystretch}{3em}
\subsection{Runtime implementation and measurement protocol}
\label{sec:results-appendix}

The WaveRT runtime sweep uses one NVLink-connected node with eight H200 141-GB GPUs, $832\times480$ resolution video, and three latent frames per chunk. We implement WaveRT using Python 3.12, PyTorch 2.11.0+cu128, and SGLang Diffusion.

For the WaveRT measurements in Tabs.~\ref{tab:runtime-core} and~\ref{tab:transport-ablation}, each repetition generates 399 latent frames and 1593 decoded frames after a 399-latent-frame warmup. System FPS is 1593 divided by the median elapsed time across three fixed repetitions. Tick p50 pools steady-state ticks across those repetitions, and steady FPS is $12000/\text{tick p50 (ms)}$, using 12 decoded-equivalent frames per steady chunk. DiT timing covers pipeline fill and drain and may be affected by concurrent VAE decoding, while E2E timing additionally includes the remaining time until VAE completion. Loading and file encoding are excluded. The BF16 configuration uses FA3 self-attention. Both Sage configurations use SageAttention for self-attention and FA3 for cross-attention, with W8A8 linear computation enabled in the Sage+W8A8 configuration.

All new 1.3B and 14B system tests use dummy DiT weights and a real-weight VAE to evaluate execution cost, not generation quality. The 14B tests use the standard Wan2.1-T2V-14B architecture.

\subsection{Throughput accounting and repeatability}

Tab.~\ref{tab:runtime-core} reports steady, complete-request DiT, and E2E rates together for each runtime configuration. Fill/drain and decoder service explain why these rates are not interchangeable. In particular, the lowest 1.3B tick p50 (WaveRT + Sage + W8A8) does not yield the highest E2E FPS.

\paragraph{Output-path limits in 5-step generation.}
Under the 8-GPU budget, 5-step WaveRT assigns five GPUs to denoising, one to Store, and two to a two-stage VAE pipeline (PP2), compared with three VAE stages in the 4-step setup. In the 5-step system measurements, BF16 and SageAttention achieve 104.82 and 104.80 DiT FPS, while E2E rates are 96.46 and 96.48 FPS, respectively (Tabs.~\ref{tab:quality-main} and~\ref{tab:runtime-core}). VAE concurrency is a plausible limiter under the smaller decoder budget, but system DiT timing can itself include concurrent decoder effects. These aggregate measurements do not isolate backpressure or establish a large hidden SageAttention speedup (Sec.~\ref{sec:vae-limitations}).

The same-graph transport ablation (Fig.~\ref{fig:systems-performance}b and Tab.~\ref{tab:transport-ablation}) compares Sync with asynchronous delivery. Tab.~\ref{tab:repeat-ranges} reports the minimum and maximum per-run tick p50 across three repetitions on the same node for causal 4-step, 8-GPU WaveRT. These ranges summarize the same experimental runs as the matching configurations in Tabs.~\ref{tab:runtime-core} and~\ref{tab:transport-ablation}, while Tab.~\ref{tab:cp-dit} reports a separate DiT-only sweep.

\begin{table}[htbp]
\centering\small
\caption{Minimum--maximum of per-run tick p50 (ms) across three fixed repetitions on the same node for causal 4-step, 8-GPU WaveRT with dummy DiT weights. These ranges are not confidence intervals.}
\label{tab:repeat-ranges}
\setlength{\tabcolsep}{3pt}
\resizebox{\linewidth}{!}{%
\begin{tabular}{@{}ll*{5}{c}@{}}
\toprule
\tabhead{Scale} & \tabhead{Compute} & \tabhead{Sync} & \tabhead{Overlap} & \tabhead{One-sided} & \tabhead{Staggered} & \tabhead{Paged} \\
\midrule
1.3B & BF16 & 91.946--92.009 & 85.659--85.763 & 81.589--81.744 & 81.842--82.107 & 78.851--78.967 \\
1.3B & Sage & 91.390--91.462 & 84.035--84.168 & 80.933--80.953 & 80.806--81.235 & 78.290--78.400 \\
1.3B & Sage+W8A8 & 88.579--88.781 & 81.505--81.634 & 77.975--78.099 & 77.895--78.291 & 75.769--76.284 \\
\midrule
14B & BF16 & 463.785--464.761 & 449.001--450.082 & 425.517--426.670 & 424.276--425.844 & 413.094--414.374 \\
14B & Sage & 466.946--468.320 & 439.004--440.453 & 425.819--426.635 & 425.420--425.753 & 413.711--414.304 \\
14B & Sage+W8A8 & 407.121--407.384 & 379.268--380.479 & 365.482--366.516 & 366.552--367.034 & 354.189--354.720 \\
\bottomrule
\end{tabular}%
}
\end{table}

\subsection{Mechanism diagnostics}

\begin{table}[htbp]
\centering\small
\caption{Diagnostic-only component profiles of stage service, K/V transfer, receiver-side assembly, and VAE partitioning, using three-run means. Each row reports a component timing within its stated measurement scope. Because these components can overlap during execution, their durations cannot be summed to obtain E2E latency. The producer range spans rankwise means.}
\label{tab:mechanism-diagnostics}
\begin{tabularx}{\linewidth}{@{}lXc@{}}
\toprule
\tabhead{Diagnostic} & \tabhead{Measurement} & \tabhead{Value} \\
\midrule
14B critical consumer & r0 forward interval & 414.23 ms \\
14B producers & r1/r2/r3/store forward interval & 370.50--379.37 ms \\
Aggregate rank-time difference & Critical minus mean of other ranks & 38.32 ms \\
BF16 K/V communication & Rank 0 collective time summed over layers & 58.45 ms \\
FP8 K/V communication & Payload and scale transfer, excluding conversion & 31.23 ms \\
Copy-engine P2P & 95,846,400-byte IPC copy & 391.49 GB/s \\
Context assembly & One-sided to paged & 19.02 to 5.91 ms \\
VAE FLOP-based split & Maximum isolated segment per chunk & 115.76 ms \\
VAE time-based split & Three isolated segments per chunk & 73.15 / 60.91 / 80.89 ms \\
\bottomrule
\end{tabularx}
\end{table}

Tab.~\ref{tab:mechanism-diagnostics} reports component-level profiles. The 38.32-ms difference compares aggregate forward intervals of the critical rank and the mean of other ranks, not per-layer readiness slack (Appendix~\ref{sec:timing-contract}). BF16 and FP8 collective diagnostics are 58.45 and 31.23 ms per tick, excluding FP8 conversion, while one-sided and paged context assembly measure 19.02 and 5.91 ms per tick in separate profiling runs. The P2P rate is in decimal GB/s. These components can overlap with computation, so their E2E impact depends on exposed waits rather than the sum of their durations. The VAE split diagnostics report 115.76 ms for the slowest FLOP-based segment and 73.15/60.91/80.89 ms for time-based segments. Because the VAE segments are profiled sequentially on one GPU without inter-GPU transfers or queue waits, these measurements characterize isolated segment latency rather than online VAE capacity.

\paragraph{Transport correctness.}
Earlier checks of the 1.3B, 4-step configuration found that synchronized, one-sided, and paged execution of the same WF graph produced bit-identical latent outputs (\texttt{torch.equal=True}, zero maximum absolute difference). Same-graph scheduling equivalence is established in Appendix~\ref{sec:semantic-correctness}.

\subsection{Exposed communication, synchronization, and payload size}
\label{sec:kv-bandwidth}

WaveRT targets time exposed on the consumer's critical path, not only bytes on the link. Early publication overlaps delivery with computation. One-sided writes replace matching collective coordination with point-of-use readiness checks, and paged placement avoids rebuilding remote context. Together, these changes reduce exposed communication, synchronization, and destination-assembly costs without eliminating data dependencies. At fixed BF16 computation, the 4-step 14B Sync/Overlap/One-sided/Paged tick p50 values are 464.270, 449.689, 425.907, and 413.795\,ms (Tab.~\ref{tab:transport-ablation}).

Payload size determines how much delivery must fit under computation. One chunk sent to one consumer at one layer carries $m_e(b)=2N_Cn_{\mathrm{head}}d_{\mathrm{head}}b$ bytes, with $b$ bytes per element. Larger K/V width increases each payload, while more layers and destinations increase round traffic (Appendix~\ref{sec:storage-communication}). Fixed hardware has unchanged peak link capacity, but effective bandwidth varies with topology and contention. Taking unquantized K/V readiness as the reference, a simple usable-delivery model is
\begin{equation}
 \widetilde\tau_e(b)\approx
 t_e^{\mathrm{quant}}+\alpha_e+
 \frac{m_e(b)}{B_e^{\mathrm{eff}}}
 +t_e^{\mathrm{dequant}},
 \label{eq:kv-delivery-budget}
\end{equation}
where $\alpha_e$ covers queueing and signaling, and $B_e^{\mathrm{eff}}$ is achieved payload bandwidth. BF16 requires no conversion, whereas FP8 halves payload bytes, excluding metadata, at the cost of conversion. With unchanged local timing, a previously exposed dependency can become hidden if
\begin{equation}
 \widetilde\tau_e(1)\leq\Delta_e<\widetilde\tau_e(2).
 \label{eq:kv-precision-threshold}
\end{equation}
Every required edge, including store, must meet its deadline: local \emph{attention} readiness, not projection start. Quantization can change execution timing, so actual slack must be measured per configuration.

The component profiles in Tab.~\ref{tab:mechanism-diagnostics} illustrate reduced transfer volume with FP8 K/V and reduced receiver-side assembly with paged placement. FP8 conversion can offset transfer savings under SageAttention. In the paged backend sweep (Fig.~\ref{fig:systems-performance}e,f), K/V remains unquantized, while W8A8 changes local DiT linear computation. These configurations vary the local compute backend while retaining the same K/V transport precision.

\FloatBarrier
\endgroup

\section{Comparison of Execution Designs}
\label{sec:schedules-appendix}

\subsection{Generation schedules and conditioning}

Similar chunk--stage diagrams need not compute the same model. Tab.~\ref{tab:schedule-comparison} distinguishes the state supplied to attention from the placement of work. All methods retain full within-chunk attention, but their weights, denoising schedules, and cache rules are method-specific.

\begin{table}[htbp]
\centering\small
\caption{Conditioning and exchanged state in different generation schedules. Live K/V is produced by active chunks at the current Transformer layer.}
\label{tab:schedule-comparison}
\begin{tabularx}{\linewidth}{@{}l*{3}{>{\raggedright\arraybackslash}X}@{}}
\toprule
\tabhead{Schedule} & \tabhead{Context state} & \tabhead{Across-stage transfer} & \tabhead{Execution structure} \\
\midrule
Self Forcing & Completed clean-history K/V & None & Sequential chunks \\
Live Avatar & Fixed-timestep stage-local K/V snapshots & Updated latent & Stage pipeline \\
HiAR & Refreshed output-timestep K/V within each hierarchy & Updated latent & Causal hierarchy pipeline \\
Rolling Forcing & Clean history and bidirectional live K/V & All active K/V if distributed & Joint mixed-noise denoising \\
Wave Forcing & Clean history and causal live K/V & Cleaner-to-noisier K/V and latent & Directed wavefront \\
\bottomrule
\end{tabularx}
\end{table}

\paragraph{Sequential and joint-chunk generation.}
Self Forcing~\citep{huang2025selfforcing} completes all updates of one chunk before starting the next, reusing clean-history K/V. Its bounded history cache can evict the initial chunk, whereas RF retains it as an anchor. RF~\citep{liu2025rollingforcing} advances consecutive mixed-noise chunks jointly, emitting one clean chunk per joint call after fill. Distributed RF retains every active chunk's corresponding-layer K/V dependencies in both temporal directions. Communication overlap is still possible, but every consumer must respect those inputs. Both RF and WF are DAGs when unrolled over layers, so WF is distinguished by cleaner-only visibility, not merely by a topological order.

\paragraph{Stage-local and matched-noise pipelines.}
Live Avatar~\citep{huang2025liveavatar} assigns a fixed denoising timestep to each process. Earlier stages retain their original K/V snapshots as a chunk progresses, and only latents move across stages. HiAR~\citep{zou2026hiar} instead builds a causal hierarchy: each level uses freshly produced K/V from earlier chunks at its output timestep $\nu_{s+1}$. The cache write for chunk $i$ can be fused with denoising chunk $i+1$. Its pipelined levels retain independent matched-noise caches and exchange latents, not layer-wise K/V between processes. This refreshed output-state context distinguishes HiAR from fixed-timestep snapshots.

\paragraph{Wavefront execution.}
WF preserves current-layer live K/V from all legal cleaner chunks and store, and WP publishes this K/V across full-model replicas. Thus an anti-diagonal schedule alone does not imply WF conditioning. Its causal mask, cache rules, and training establish the model, while publication and readiness checks implement it. Appendix~\ref{sec:execution-cost} gives the cadence and resource accounting without equating steady output rate with first-output latency.
\subsection{Partitioning a forward versus distributing stages}
\label{sec:parallelism-analysis}

\begin{table}[htbp]
\centering
\small
\caption{Resource-partitioning strategies, communication patterns, and constraints. TP and CP preserve the chosen graph, while WP schedules the fixed WF graph.}
\label{tab:parallel-strategies}
\begin{tabularx}{\textwidth}{l*{3}{>{\raggedright\arraybackslash}X}}
\toprule
\tabhead{Method} & \tabhead{Partitioned object} & \tabhead{Main communication} & \tabhead{Main constraint} \\
\midrule
Tensor parallelism & Hidden/head dimensions & Per-operator collectives & Small local GEMMs at high degree \\
Ulysses CP & Tokens to attention heads & Two All-to-All phases & Head divisibility and per-layer collectives \\
Ring Attention & K/V sequence blocks & Multi-round K/V circulation & Enough tile compute to hide P2P \\
Chunk-owned sync & Active query chunks & Layer K/V All-Gather & Global layer barrier \\
Wave Parallelism & Chunk--stage tasks & One-way K/V and latent & Full-model replicas and sufficient overlap slack \\
\bottomrule
\end{tabularx}
\end{table}

Tensor and context parallelism partition one forward while retaining its conditioning graph. Ulysses~\citep{jacobs2023deepspeedulysses} limits CP degree by attention-head divisibility, whereas Ring Attention~\citep{liu2023ringattention} supports head-independent degrees through ordered K/V circulation. Synchronized chunk ownership keeps local queries and all their heads intact but retains Eq.~\eqref{eq:sync-pool}'s layer collective. WP instead relies on the WF graph and sufficient producer lead, at the cost of full-model replication.

\paragraph{Comparison setup.}
The three-backend comparison in Fig.~\ref{fig:systems-performance}c--f and Tab.~\ref{tab:cp-dit} matches parameter count, backend tier, GPU count, and denoising-step count across methods within each model scale: a 5-step, 6-GPU setup for 1.3B--9B and a 4-step, 5-GPU setup for 14B. These rates are mean per-run DiT FPS with VAE disabled, whereas Fig.~\ref{fig:systems-performance}a reports E2E throughput for the 4-step, 8-GPU setup. Over the same rolling window, CP uses RF's full active-window attention, whereas WP uses WF's block-causal mask. The comparison therefore evaluates the combined effect of execution strategy and attention visibility. Tab.~\ref{tab:parallel-strategies} summarizes how each strategy partitions work, communicates dependencies, and constrains resource allocation.

\paragraph{Model Configuration.}
The 1.3B and 14B models retain the standard Wan2.1 architectures. They share the same Wan Transformer block design but differ in the number of blocks, attention heads, hidden dimensions, and FFN dimensions. For the parameter-scale comparison in Tab.~\ref{tab:cp-dit}, we adjust these dimensions and the number of blocks to construct the intermediate 3B, 5B, 7B, and 9B architectures. Their attention-head counts are divisible by six to support six-way Ulysses parallelism. The standard 14B architecture has 40 attention heads and supports the five-way configuration used in our comparison. Tab.~\ref{tab:synthetic-arch} lists all six configurations.

All six scales use dummy weights in the execution tests, including the standard Wan2.1-T2V-1.3B and 14B architectures. These tests measure execution performance separately from the trained 1.3B quality checkpoints. We do not train a 14B Wave Forcing model because the required computation exceeds our available resources.

\paragraph{Repetition protocol.}
Unmarked entries in Tab.~\ref{tab:cp-dit} report the arithmetic mean of three fixed runs. The $\dagger$-marked entries are supplemental diagnostics collected with throughput-based stopping.

\begin{table}[htbp]
\centering
\small
\setlength{\tabcolsep}{8pt}
\caption{Model configurations for the parameter-scale comparison.
The upper group contains the standard Wan2.1 architectures,
and the lower group contains the synthetic architectures.}
\label{tab:synthetic-arch}
\begin{tabular}{@{}l*{4}{c}@{}}
\toprule
\tabhead{Parameters} & \tabhead{Layers} & \tabhead{Heads} & \tabhead{Hidden dim.} & \tabhead{FFN dim.} \\
\midrule
1.3B & 30 & 12 & 1536 & 8960  \\
14B  & 40 & 40 & 5120 & 13824 \\
\midrule
3B   & 32 & 18 & 2304 & 10752 \\
5B   & 30 & 24 & 3072 & 14336 \\
7B   & 29 & 30 & 3840 & 15360 \\
9B   & 26 & 36 & 4608 & 18432 \\
\bottomrule
\end{tabular}
\end{table}

\FloatBarrier

\clearpage
\section{Extended Future Work}
\label{sec:broader-connections}

This appendix develops the interaction, serving, and world-model extensions in Sec.~\ref{sec:future-work}, following the training and execution limits in Sec.~\ref{sec:discussion}. These extensions have not been evaluated with WF.

\paragraph{Online interaction and condition changes.}
LongLive~\citep{yang2025longlive} uses KV recaching to align visual context with an updated prompt; EditStream and LPM~\citep{zhou2026editstream,zeng2026lpm} explore interactive editing and character performance. A WF extension must specify which prompt applies to each active chunk, which cached states need refreshing, and whether outstanding work is completed or discarded. The fixed-context equivalence argument in Appendix~\ref{sec:semantic-correctness} does not establish correctness across condition changes. Sustained FPS and interaction delay are distinct: evaluation should measure when an input first changes the displayed video, as well as adherence and continuity.

\paragraph{Serving under streaming deadlines.}
StreamDiffusion's image-stream batching~\citep{kodaira2025streamdiffusion} and StreamDiffusionV2's video-stream scheduling~\citep{feng2026streamdiffusionv2} motivate deadline-aware batching around WF. SlackServe~\citep{xia2026slackserve} relates scheduling and quality choices to buffered playback time, while PhysisServe's technical blog~\citep{physis2026physisserve} describes pipelining over transformer blocks and denoising steps, together with GPU time sharing. Future WF serving should study admission control, stream priorities, resource allocation, and finer-grained placement under its live-context dependencies. Changing sampling budgets also changes WP's available stage parallelism (Sec.~\ref{sec:discussion}); adopting another system's batching or state-reuse rules would therefore require checking WF's conditioning graph. Startup delay, output jitter, stall rate, and fairness should accompany throughput and quality measurements.

\paragraph{Learning controlled world evolution.}
One route extends video prediction with structured controls. minWM and ForgeWM~\citep{zhao2026minwm,li2026forgewm} study controllable few-step video generation; SolarWM~\citep{huang2026solarwm} emphasizes data standardization and causal adaptation; SANA-WM~\citep{zhu2026sanawm} explores hybrid linear attention for long-horizon generation; and OneWM~\citep{su2026onewm} jointly learns world prediction and action generation. Integrating WF would require specifying when new camera or action inputs become available to in-flight chunks, validating state dependencies for each architecture, and assessing control fidelity, prediction consistency, and policy performance in closed-loop tasks.

\paragraph{Executable world state and generative rendering.}
A complementary route delegates persistent state, rules, and dynamics to code or an engine. Code World Model and Matrix-Game 4.0~\citep{chen2026codeworldmodel,zhang2026matrixgame4} connect such state to video generation through visual proxies. StreamRender-H3~\citep{streamrenderh3_2026} and SolarRenderer~\citep{fang2026solarrenderer} further motivate structure-conditioned streaming rendering. Here WF could supply an execution foundation for a renderer trained to follow incoming structure, rather than taking responsibility for the engine's dynamics. This would require aligned structure/video training and validation of temporal conditioning: StreamRender-H3 uses bidirectional attention within a streaming window, whereas SolarRenderer includes a block-causal variant. Neither establishes compatibility with WF's wavefront. Evaluation should check whether rendered observations preserve the underlying state and expose action effects quickly enough for feedback.

\paragraph{Understanding-assisted quality evaluation.}
Following the metric limits in Sec.~\ref{sec:discussion}, a future evaluator could combine local defect checks with long-horizon, instruction-aware judgments. StreamEval~\citep{streameval2026} evaluates timed requests and retained state alongside visual quality. StreamMeCo~\citep{wang2026streammeco} addresses efficient streaming video understanding, while MemForest~\citep{wang2026memforest} provides event-centric memory compression and retrieval. These mechanisms could help a video-understanding judge retrieve temporal evidence about entities, actions, and failures. Adapting them to quality judging requires human-calibrated validation: we should report supporting timestamps and unscorable intervals, assess agreement with human annotations, and test whether compressed memory loses brief defects or earlier facts. The goal is to complement our targeted diagnostics with broader, auditable quality assessment.

\end{document}